\documentclass[lettersize,journal]{IEEEtran}
\usepackage{amsmath,amsfonts}
\usepackage{algorithmic}
\usepackage{algorithm}
\usepackage{array}
\usepackage[caption=false,font=normalsize,labelfont=sf,textfont=sf]{subfig}
\usepackage{textcomp}
\usepackage{stfloats}
\usepackage{url}
\usepackage{verbatim}
\usepackage{graphicx}
\usepackage[table]{xcolor}
\usepackage{cite}
\graphicspath{ {figures/} }
\begin{document}

\title{Inverse Design of Electromagnetic Structures via Local Geometry Parameterization and Latent Space Optimization}

\author{Erez Zolkov,~\IEEEmembership{Member,~IEEE,} Vinay Chenna,~\IEEEmembership{Member,~IEEE,} Hossein Hashemi,~\IEEEmembership{Fellow,~IEEE}
}



\maketitle

\begin{abstract}
Conventional optimization of integrated electromagnetic (EM)
structures is limited by the geometric flexibility of predefined
topologies, whereas pixel-based inverse design explores a much larger
design space at the expense of numerous EM simulations. This work
presents a two-step inverse-design framework that combines local
geometry parameterization with latent-space optimization. First, a
constrained geometry generator enables independent variation of
conductor widths, positions, metal layers, and surrounding ground
structures while enforcing electrical connectivity and geometric
constraints. Second, an autoencoder trained on generated layouts
without EM simulation learns a continuous representation of these
geometric variations. Optimization is performed in this latent space,
with each candidate reconstructed into a valid multilayer layout and
evaluated through full-wave EM simulation and circuit-level
optimization. The proposed approach is compared with conventional
parameter optimization, direct optimization of the local geometry
parameters, and pixel-based direct binary search. For a
quadrature-coupler design, latent-space optimization reaches a
specified objective value with approximately 3.9 times fewer EM
simulations than direct local-parameter optimization and 45 times
fewer than direct binary search. Three broadband quadrature couplers
were fabricated in 65-nm CMOS for the 10--20, 25--50, and 30--60~GHz
bands, achieving measured insertion losses below 1.5~dB, amplitude
imbalances below 1~dB, and phase imbalances below $2^\circ$, in active
areas of 0.16, 0.067, and 0.044~mm$^2$, respectively.
\end{abstract}


\section{Introduction}
\begin{figure}[t!]
	\centering
	\subfloat[]{{\includegraphics[width=\columnwidth]{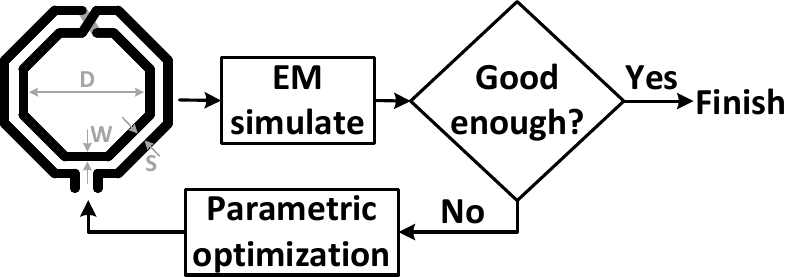}\label{fig:algo_1st}}}\\
	\subfloat[]{{\includegraphics[width=\columnwidth]{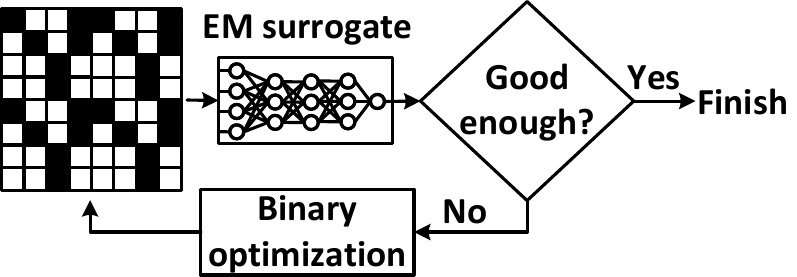}\label{fig:algo_2nd}}}\\
	\subfloat[]{{\includegraphics[width=\columnwidth]{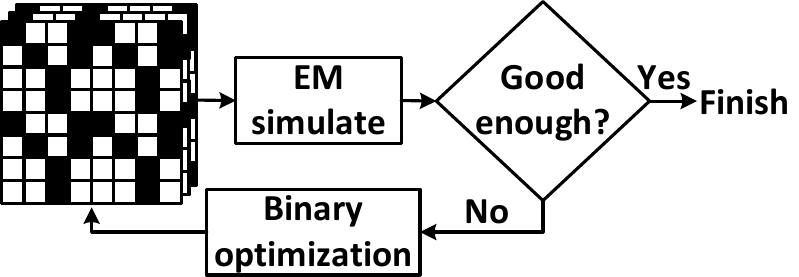}\label{fig:algo_3rd}}}\\
	\subfloat[]{{\includegraphics[width=\columnwidth]{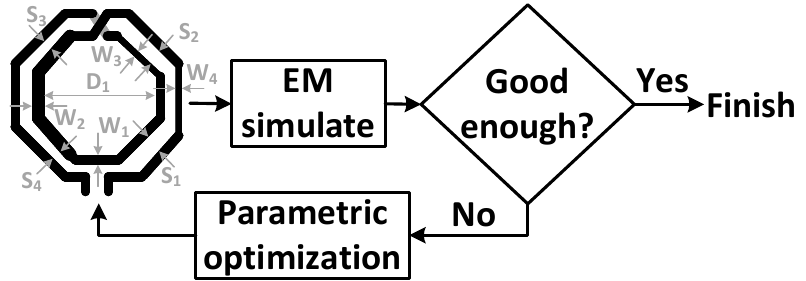}\label{fig:algo_4th}}}
	\caption{Representative algorithmic workflows: (a) parametric EM
	optimization of a textbook topology with global parameters,
	(b) surrogate-assisted optimization, (c) binary
	optimization, which flips individual pixels, and (d) the proposed
	local parameterization, in which many local knobs shape the
	structure.}
	\label{fig:algorithm_overview}
\end{figure}

\IEEEPARstart{T}{he} performance of radio-frequency and
millimeter-wave integrated circuits depends on the
electromagnetic (EM) structures used for impedance matching,
filtering, power combining, and signal coupling. These structures are
traditionally designed by selecting a known topology, such as a spiral
inductor, transformer, or transmission-line network, and iteratively
adjusting its geometric parameters with EM simulations
(Fig.~\ref{fig:algo_1st}). Global optimizers and surrogate models
accelerate this
search~\cite{bandler2004space,liu2006integrated,liu2011synthesis,gregory2011fast,koziel2015rapid},
but the parameters remain global dimensions of one topology: a single
trace width, spacing, or radius applies to the whole structure, the
metal layer of each conductor, and with it the coupling mechanism, is
fixed, and the ground is not a design variable. The designs are
therefore limited by the topology and the designer's experience, and
each added parameter makes the optimization slower.

Inverse design instead determines the geometry itself by numerical
optimization~\cite{chenna2026pixelated,molesky2018inverse}. One
approach discretizes the metal layers into a grid of pixels and
optimizes the presence of metal at each
location~\cite{john1996evolutionary,jin2005parallel,lee2023pixelated,karahan2024inverse}.
This expands the realizable geometries, which can give wider bandwidth
or lower loss than conventional designs, but makes the
optimization hard for two reasons. First, the number of possible
layouts grows exponentially with the number of pixels and metal
layers. Second, arbitrary pixel changes can violate connectivity, via
placement, and fabrication constraints, particularly in multilayer
structures.

Several approaches address these challenges. Gradient-based topology
optimization~\cite{hassan2014topology,vercruysse2019analytical}
explores high-dimensional geometries efficiently when EM sensitivities
are available, but in commercial IC design environments it is limited
by access to solver derivatives and by discrete fabrication
constraints~\cite{chenna2026pixelated}. Gradient-free methods,
including genetic algorithms~\cite{john1996evolutionary}, particle
swarm optimization~\cite{jin2005parallel}, and memetic
algorithms~\cite{capek2023optimal}, operate directly on binary
geometries but typically require many EM evaluations, even when
precomputed numerical Green's functions reduce the cost of
each~\cite{zheng2023ultrafast}. More recently, machine-learning
surrogates predict the EM response of candidate geometries to reduce
the cost of optimization
(Fig.~\ref{fig:algo_2nd})~\cite{liu2006integrated,karahan2023deep,karahan2024inverse,he2025motif,wei2023transformer,chae2024pulserf}.
Training them, however, requires EM simulation data, and their
accuracy depends on how well the dataset represents the geometries
met during optimization.

These limitations are worse in multilayer structures. Direct binary
search (DBS) and binary particle swarm optimization have been applied
to multilayer on-chip passive networks with full-wave simulation in
the loop~\cite{chenna2026pixelated} (Fig.~\ref{fig:algo_3rd}). With
three metal and two via layers, each grid location admits 13 valid
configurations, so an $M\times N$ template spans $13^{MN}$ layouts.
DBS found competitive designs within 250--500 EM simulations, but each
design took 6--10~hours, the results depended on ad hoc
hyperparameters~\cite{wolpert1997no}, and independent pixel flips had
to be repaired to preserve connectivity~\cite{chenna2026pixelated}.
Multilayer EM simulations also take 10--20 times longer than
single-layer ones of the same area~\cite{chenna2026pixelated}, making
surrogate training data correspondingly more expensive.

The two approaches thus sit at opposite extremes: a few global
parameters that always give a valid but conventional structure, or
many independent pixels that can express any structure but rarely a
valid one. This work takes an intermediate path in two steps.

The first step is a parameterization that is local but always valid
(Fig.~\ref{fig:algo_4th}). Instead of the few global dimensions of
Fig.~\ref{fig:algo_1st}, a constrained generator describes the layout
by many knobs, each acting on one part of the structure
(Section~\ref{sec:generation}). Coarse knobs set the outline, the
ground patches around the conductors, and the metal of every leg of
each net, so that lateral (side-by-side) and broadside (stacked)
coupling can be mixed along one winding; fine knobs set the width and
position of every sub-segment. The constrained generator produces a broad family of layouts beyond those accessible through the global parameters, while geometric validation and regeneration ensure that only valid layouts are evaluated. Searching the knobs with full-wave EM simulation already
outperforms both a search on the global parameters, which stalls, and
the pixel search (Section~\ref{subsec:qc_ablations}).

The second step improves this search with a learned representation.
An explicit parameterization must write every variation as a knob: the
QC of Section~\ref{sec:experimental_results} needs 247 knobs, many of
them discrete, which interact strongly and carry no notion of where in
the layout they act. We therefore train an
autoencoder (AE)~\cite{hinton2006reducing}, without EM simulation, on
layouts from a constrained generator that produces the same variations
at random, through random routing and widening of the traces and
random ground patches. The search then runs in the continuous latent
space of the AE, and every candidate is evaluated by full-wave EM
simulation. The variations need no hand-defined knobs: the latent
vector carries them, and the decoder maps them to the layout. Because
the latent space keeps the spatial arrangement of the layout, a small
latent step shifts or widens a whole trace. For the QC, the latent
space has about as many variables as the knob space, but they are
continuous and spatially correlated, whereas the knobs mix discrete and
continuous variables. Such latent space
optimization (LSO) was introduced for molecular
design~\cite{gomez2018automatic,tripp2020sample}. For integrated EM
structures, every decoded candidate must satisfy hard connectivity and
via constraints, which a neural decoder does not guarantee, and the
geometry must be optimized together with the lumped circuit elements
that load it. Here, the decoder guides the generator to rebuild every
candidate as a valid layout, which is scored with its lumped elements
tuned. For a quadrature-coupler design, latent-space optimization reaches a specified objective value with 3.9 times fewer EM simulations than direct local-parameter optimization (median of five runs) and 45 times fewer than a representative direct binary search run.(Section~\ref{subsec:qc_ablations}).

The main contributions of this work are as follows.
\begin{enumerate}
\item A constrained parametric generator with local knobs on each
section of every net and on the surrounding ground, so that the metal,
width, and position, and with them the type of coupling, can change
along a winding, while every generated layout remains valid.
\item A geometry-only learned representation of the generated
layouts, requiring no EM simulation, with a guided reconstruction in
which the decoder output and a learned knob predictor drive the
generator, so every latent point maps to a valid multilayer layout.
\item A physics-based latent search that combines covariance matrix
adaptation evolution strategy (CMA-ES) with nested tuning of the
lumped circuit elements and ensemble-Kalman proposals that exploit the
per-frequency specification residuals.
\item An experimental demonstration with three fabricated broadband
quadrature couplers (QCs), and a comparison with a common EM solver,
circuit tuning, and loss that separates the two steps: the local knobs against the classic width,
spacing, and diameter and against DBS of the pixels, and the latent
search against the local knobs.
\end{enumerate}

Section~\ref{sec:generation} describes the local parameterization,
Section~\ref{sec:concept} the concept of the latent search,
Section~\ref{sec:representation} the learned representation, and
Section~\ref{sec:optimization} the physics-based search.
Section~\ref{sec:experimental_results} presents the QC designs,
measurements, and comparisons, and Section~\ref{sec:conclusion}
concludes the article.

\section{Constrained Layout Generation}
\label{sec:generation}
\label{sec:general_inverse_design}

The first step describes the base layout by local knobs
(Fig.~\ref{fig:algo_4th}).

\begin{figure}[t!]
	\includegraphics[width=\columnwidth]{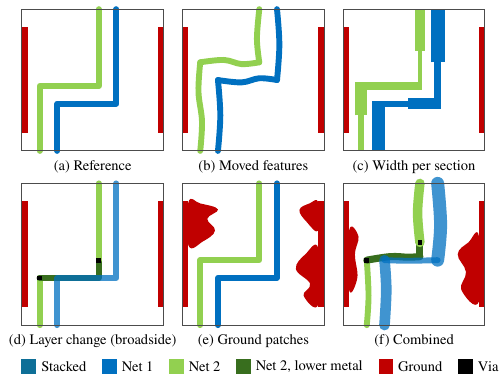}
	\centering
	\caption{Illustration of the constrained geometry generator on an
	arbitrary two-net reference: (a) reference and (b)--(f)
	perturbations that keep the ports fixed and every net connected:
	(b) moved features with random deviations of the traces,
	(c) a different width in each section, (d) a section of net~2
	moved to the lower metal, where it is stacked under net~1 (broadside
	coupling), (e) ground patches, and (f) all combined, with the
	lower-metal section of net~2 shifted partly beside net~1.}
	\label{fig:generic_examples}
\end{figure}

\subsection{Continuous Signal Geometry}
\label{subsec:general_signal_geometry}

Each signal net is a closed-form polyline from one port to the other
along the sides of a reference outline, such as the chamfered octagon
of an XFMR. It is drawn as exact polygons of a given width and metal,
with vias at every change of metal, and passed to the EM solver
without discretization. The route is divided into legs, one per side
or corner of the outline, and each leg into sub-segments that meet at
break positions.

\subsection{Coarse Knobs and Fine Perturbations}
\label{subsec:general_mask_generator}

The generator builds the polygons
$\Pi=\mathcal B_\Pi(\mathbf k_c,\mathbf k_f)$ from two levels of
knobs (Fig.~\ref{fig:generic_examples}). The coarse knobs
$\mathbf k_c$ act on whole legs, tracks, or nets: the position and
chamfers of the outline, the metal of each leg and crossover, a
common spread of the tracks, a per-track offset, and common and
per-net width increments. A leg lies on the top metal, on the lower
metal stacked under the facing track of the other net, or on the
lower metal at its own position, so the coupling changes between
lateral (side by side) and broadside (stacked) from leg to leg
(Fig.~\ref{fig:generic_examples}(d)).

The fine knobs $\mathbf k_f$ act locally. Each sub-segment has its own
width and inset from the reference side, and sub-segments with
different insets meet in a $45^\circ$ jog, so the track position,
coupling gap, and taper vary along a leg
(Fig.~\ref{fig:generic_examples}(b) and (c)). The break positions, the
position and length of the crossover jogs, and the width, metal, and
via depth of the feeds are also knobs. A lower-metal leg at its own
inset can lie partly beside the other net, giving a coupling between
broadside and lateral (Fig.~\ref{fig:generic_examples}(f)). The
effective inset of a sub-segment is its fine inset, scaled by the
common spread and shifted by its track offset, and its effective width
is its fine width plus the common and per-net increments, so one
coarse knob can move a whole track or widen a whole net.

Ground patches are added on the pixel grid around the signal. Their
knobs, part of $\mathbf k_c$, set the presence and density of the
patches, their depth profile along each edge, and the fractions of
ground on the lower metal and on vias
(Fig.~\ref{fig:generic_examples}(e)). The patches are trimmed to the
minimum spacing from the signal polygons.

\subsection{Validity}
\label{subsec:general_validity}

Validity is checked on the exact geometry. A layout is accepted only
if every leg advances along its side without folding back, all metal
stays on the die, different nets keep the minimum spacing on each
metal, and no two distant parts of one net come close enough to weld.
A failing layout is redrawn, so every accepted layout is drawable,
with connected nets and landed vias.

The knob vector $(\mathbf k_c,\mathbf k_f)$ can be searched directly,
as in Section~\ref{subsec:qc_ablations}.

\section{Latent Space Inverse Design: Concept and Architecture}
\label{sec:concept}

\begin{figure}[t!]
	\includegraphics[width=\columnwidth]{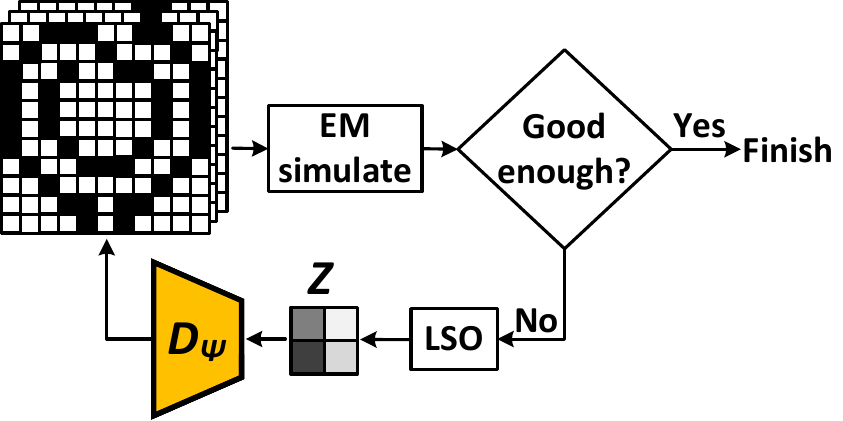}
	\centering
	\caption{The proposed LSO workflow.  The optimizer searches the continuous latent variables $\mathbf z$. The decoder $D_\psi$ provides spatial guidance to a constrained geometry generator that reconstructs valid multilayer layouts for EM evaluation.}
	\label{fig:LSO_simple}
\end{figure}

\begin{figure}[t!]
	\includegraphics[width=\columnwidth]{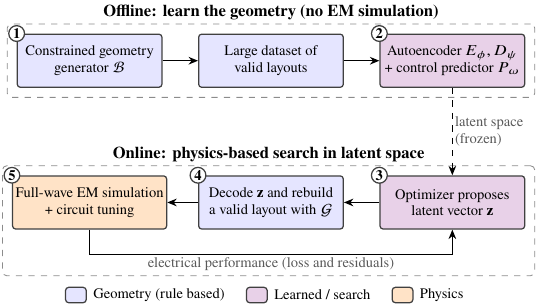}
	\centering
	\caption{Concept of the proposed latent space inverse design. Offline,
	a constrained generator produces valid layouts and an AE learns a
	compact representation of them, without EM simulation (steps 1--2).
	Online, an optimizer searches the latent space, each candidate is
	rebuilt into a valid layout, and its performance is evaluated by
	full-wave EM simulation with circuit tuning (steps 3--5).}
	\label{fig:concept}
\end{figure}
The second step replaces direct optimization of the explicit geometric knobs with optimization of a learned representation of the generated layouts. It moves from the polygons of Section~\ref{sec:generation} to
a pixel grid, on which the convolutional AE operates: the training
layouts, the decoder output, and every candidate are categorical masks
of the metal and via planes (Section~\ref{subsec:general_dataset}).
The masks come from a second generator, which differs from that of
Section~\ref{sec:generation} in what it controls explicitly. There,
every per-section variation, such as the width and inset of each
sub-segment and the metal of each leg, is a knob. The mask generator
has only a few structural controls, such as the positions of the main
features, a nominal width, and a layer plan; the per-section
variations come from its random routing, widening, and ground
placement. After training, the latent vector carries them.

The loop is that of binary optimization (Fig.~\ref{fig:algo_3rd})
with a different step (Fig.~\ref{fig:LSO_simple}): the optimizer
changes the latent variables $\mathbf z$, and the decoder turns them
into coordinated changes of many pixels. The representation is learned
once, offline, from geometry alone; the search runs online, with every
candidate simulated. Fig.~\ref{fig:concept} shows the five steps.

\textit{1) Generate valid layouts.} A constrained generator, run with
random structural controls, routing, widening, and ground, produces many valid
layouts cheaply (Section~\ref{subsec:general_dataset}).

\textit{2) Learn a compact representation
(Section~\ref{sec:representation}).} An AE is trained to compress each
layout into a small latent tensor and reconstruct it, which forces it
to capture what makes a layout valid, such as continuous traces and
consistent spacing. More than $10^4$ discrete pixel decisions are
replaced by a few hundred continuous variables, in which nearby points
are similar layouts. The latent tensor keeps
the spatial arrangement of the layout, although a single latent
variable can affect several features of it. A small predictor recovers from the latent vector the
structural controls that the generator needs.

\textit{3) Search in the latent space.} CMA-ES, a gradient-free
evolutionary optimizer, proposes latent vectors in place of the pixel
flips of DBS. It suits a continuous space of moderate dimension that
is accessible only through black-box evaluations. Ensemble-Kalman
proposals add candidates computed from the per-frequency
specification residuals, which the CMA-ES ranking discards.

\textit{4) Rebuild a valid layout from each candidate.} The decoder
output, a set of per-pixel class probabilities, need not form a valid
layout. The generator of step 1 therefore uses it as guidance: the
probabilities set its routing costs, and the predicted controls place
the main features. A given latent vector always yields the same
layout.

\textit{5) Evaluate with full-wave EM simulation and tune the circuit.}
Each new layout is simulated once with a full-wave EM solver. The
lumped networks that load its ports are then tuned at the circuit
level, since a geometry can only be judged with the circuit values
that suit it. The loss and the per-frequency residuals return to the
optimizer.

Machine learning here represents only the geometry, and every
performance value comes from full-wave simulation. The optimizer
therefore cannot exploit surrogate errors, as it can in
surrogate-assisted design (Fig.~\ref{fig:algo_2nd}), and the AE need
not be retrained when the specifications or the band change. Only
the reference layout, generator rules, validator, and objective are
device specific.

\section{Learning a Geometry Representation}
\label{sec:representation}

\subsection{Training Layouts}
\label{subsec:general_dataset}
\label{subsec:general_reference_mask}
\label{subsec:general_pixel_generator}

The training layouts are represented by the categorical mask
\begin{equation}
    M \in
    \{0,\ldots,K-1\}^{L\times H\times W},
    \label{eq:general_mask}
\end{equation}
where $L$ is the number of metal and via planes, $H\times W$ is the
grid, and the $K$ labels are empty space, one label per signal net,
and ground. Separate signal labels make a short between nets visible
in the mask. The reference mask $M_0$ comes with a structural contract that fixes the
layer order, the via planes, the port pixels of each net, and the
regions reserved for ground.

The masks are produced by a constrained generator
$M=\mathcal B_M(M_0,\mathbf u,\boldsymbol\zeta)$ that covers the same
kinds of layouts as the parameterization of Section~\ref{sec:generation}
but with far fewer explicit controls. Its structural controls
$\mathbf u$ set only the positions of the main features of the
reference, a nominal trace width, the metal of each net at a few
stations of its route, including broadside stacking; its random
choices $\boldsymbol\zeta$ set everything else, including the width
and position of every section and the ground. Each net is routed
between its features by an A$^*$ search~\cite{hart1968formal} whose
costs keep it near the course of the reference and whose random terms
make it deviate. The routes are then widened by random growth that
keeps the clearance between nets, so the realized width varies from
section to section. Ground is added inward from the edges of the grid,
with a random depth profile along each edge. A validator
$\mathcal V$ accepts a mask, with its port metadata $\mathcal P$, only
if every net connects its ports as one component, different nets keep
a one-pixel clearance, every via lands on metal on both sides, and no
metal floats; rejected samples are rerouted.

Repeated sampling gives a dataset of $N$ validated masks
$\{M^{(n)},\mathcal P^{(n)},\mathbf u^{(n)}\}_{n=1}^{N}$ with their
realized controls. Since no EM simulation is needed, the dataset can be
far larger than the number of layouts simulated in a search. Each mask
is converted into a one-hot tensor
$X\in\{0,1\}^{LK\times H\times W}$ for the network.

\subsection{AE Representation and Training}
\label{subsec:general_autoencoder}

\begin{figure}[t!]
	\centering
	\subfloat[]{{\includegraphics[width=0.5\textwidth]{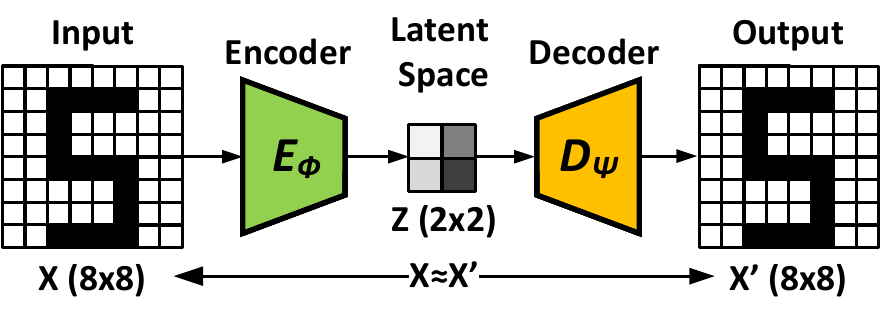}\label{fig:AE_classic}}}  \\
	\subfloat[]{{\includegraphics[width=0.5\textwidth]{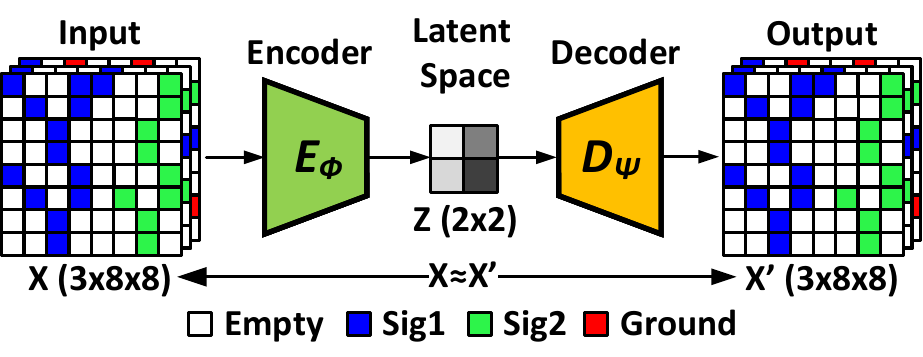}\label{fig:AE_labels}}}
	\caption{(a) Concept of an AE and (b) its application to layout representation with net labeling.}\label{fig:AEs}
\end{figure}

An AE reproduces its input through a lower-dimensional latent
code~\cite{hinton2006reducing}: an encoder maps the input to the code,
and a decoder maps the code back (Fig.~\ref{fig:AE_classic}). The
narrow bottleneck forces the code to keep the features that matter
most. Here, the input is a categorical layout in which each pixel is
empty space, ground, or one of the signal nets
(Fig.~\ref{fig:AE_labels}), and the AE maps a one-hot mask to a latent
tensor and back to categorical logits,
\begin{equation}
    \mathbf z = E_{\phi}(X)\in\mathbb R^{N_z\times H_z\times W_z},
    \qquad
    \boldsymbol\ell = D_{\psi}(\mathbf z)\in\mathbb R^{LK\times H\times W},
    \label{eq:general_encoder}
\end{equation}
where $\phi$ and $\psi$ are the trainable parameters, $N_z$ is the
number of latent channels, and $H_z\times W_z$ is the latent grid.
Both networks are convolutional, so the latent tensor keeps the
spatial arrangement of the layout.

Signal pixels are reconstructed exactly, and ground only as a soft
occupancy field. The objective also prepares
the latent space for optimization (Appendix~\ref{app:ae_objective}):
noise injected into the latent code teaches the decoder to tolerate
small latent steps, a connectivity penalty discourages disconnected
signal regions and inconsistent vias, and smoothness terms keep the
output stable under small latent perturbations. These terms make a
valid decode likely but not certain, so every candidate is rebuilt by
the constrained generator (Section~\ref{subsec:general_guidance}).
After training, the latent coordinates are standardized elementwise
with the dataset mean $\boldsymbol\mu$ and standard deviation
$\boldsymbol\sigma$, giving the search coordinates
$\widetilde{\mathbf z}$.

\subsection{Structural Prediction}
\label{subsec:general_knob_predictors}

The decoder shows where the conductors should be, but the generator
also needs its structural controls. A network $P_\omega$ estimates
them from the standardized latent vector,
$\widehat{\mathbf u}=P_\omega(\widetilde{\mathbf z})$, treating each
control as a classification target. It is trained on masks generated
with forced controls and encoded by the frozen AE, so each target is
the control actually realized (Appendix~\ref{app:predictor}).
Controls visible in the decoder output, such as the metal of a net,
are read from the logits, by comparing the class probability
mass in windows at known locations.

\section{Physics-Based Optimization in Latent Space}
\label{sec:optimization}
\label{subsec:general_execution}

\begin{figure}[t!]
	\includegraphics[width=0.5\textwidth]{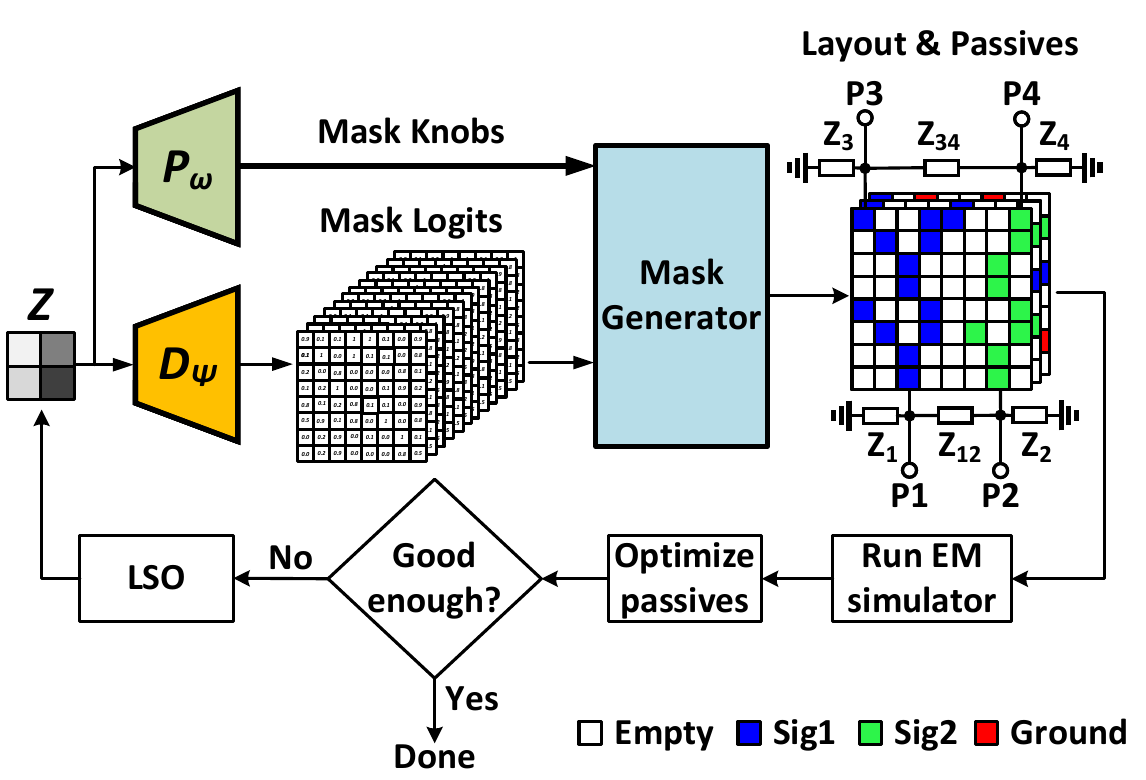}
	\centering
	\caption{LSO block diagram, shown for a
	four-port layout with three layers and four pixel classes. The
	ports P1--P4 of the pixelated layout are loaded by arbitrary
	external impedances, $Z_1$--$Z_4$ from each port to ground and
	$Z_{12}$ and $Z_{34}$ between port pairs, whose element values are
	tuned for every simulated layout.}
	\label{fig:LSO_diagram}
\end{figure}

With the AE and the predictor frozen, the design is optimized in the
latent space. Fig.~\ref{fig:LSO_diagram} shows the optimization loop.
A latent vector $\mathbf z$ is decoded into mask logits by $D_\psi$,
and $P_\omega$ predicts the structural controls. From both, the mask generator
rebuilds a valid layout. The layout is simulated, and the passives at
its ports are tuned. If the result is not yet good enough, LSO
proposes a new $\mathbf z$ and the loop repeats.

\subsection{Guided Mask Reconstruction}
\label{subsec:general_guidance}

For a standardized latent vector, the decoder logits
$\boldsymbol\ell=D_\psi(\boldsymbol\mu+\boldsymbol\sigma\odot\widetilde{\mathbf z})$
are converted into per-pixel class probabilities $p_{l,c}(y,x)$ by a
softmax over the $K$ classes, and then into routing costs
\begin{equation}
    g_{l,c}(y,x)
    =
    \operatorname{clip}\!\left(
    -\kappa\log p_{l,c}(y,x),\,g_{\min},\,g_{\max}\right),
    \label{eq:general_guidance_cost}
\end{equation}
where $l$, $c$, and $(y,x)$ index the plane, class, and pixel, and
$\kappa$ sets the guidance strength. The bounds $g_{\min}$ and
$g_{\max}$ keep confident predictions from creating impassable walls.
The guided generator $\mathcal G$ is the generator of
Section~\ref{subsec:general_pixel_generator} with these routing costs
and the controls $\widehat{\mathbf u}$. It returns a mask $\widehat M$
that passes the validator $\mathcal V$. Ground is grown where the
decoded ground probability is high, up to the decoded ground mass.
Every routing choice is seeded, so a latent vector always yields the
same mask. If construction fails, the guidance strength and then the
controls are relaxed until a valid mask is found. The map from latent
vector to mask is thus valid everywhere but piecewise constant, since
rounded controls and alternative routes produce plateaus and jumps,
which favors a gradient-free optimizer.

\subsection{EM Evaluation and Circuit Tuning}
\label{subsec:general_electrical_objective}

For a search vector $\mathbf x=\operatorname{vec}(\widetilde{\mathbf z})$,
the reconstructed mask $M(\mathbf x)$ is converted into a physical
layout and simulated once with a full-wave EM solver as a multiport
structure. As in Fig.~\ref{fig:LSO_diagram}, its ports are loaded by
impedances from each port to ground and between port pairs, each a
lumped network whose form is set by the application. The response $\mathbf S(f;\mathbf x,\boldsymbol\theta)$ is the circuit
simulation of the EM-extracted structure together with this network,
whose element values form the circuit variables $\boldsymbol\theta$.
Because the EM simulation does not depend on $\boldsymbol\theta$,
the circuit variables can be retuned without further EM simulation.

Each specification is turned into a normalized violation per
frequency: two-sided targets use the absolute error, and one-sided
limits use $\max(0,\cdot)$. Violations are aggregated with the log-sum-exp (LSE)
smooth maximum
\begin{equation}
    \operatorname{LSE}_{\alpha}(\mathbf e)
    =
    \frac{1}{\alpha}
    \log\sum_k\exp(\alpha e_k),
    \label{eq:general_softmax_loss}
\end{equation}
first over frequency and then across metric categories, so that the
loss approximates the worst violation and depends on all of them. Each
geometry is ranked by its loss after circuit tuning:
\begin{equation}
    F(\mathbf x)
    =
    \min_{\boldsymbol\theta\in\Theta}
    \mathcal L\!\left[
        \mathbf S(f;\mathbf x,\boldsymbol\theta)
    \right],
    \label{eq:general_bilevel}
\end{equation}
where $\mathcal L$ is the aggregated loss and $\Theta$ is the box of
admissible circuit values. The inner minimization is a
Nelder--Mead (NM)~\cite{nelder1965simplex} search with a fixed budget
of circuit simulations, warm-started from the values of nearby,
previously tuned designs. It is cheap compared with the EM simulation,
which runs once per distinct mask.

\subsection{CMA-ES Search in the Learned Representation}
\label{subsec:general_cmaes}

CMA-ES~\cite{hansen2001completely,hansen2016cma} searches the
standardized latent vector $\mathbf x$, of dimension
$D=N_zH_zW_z$. CMA-ES has proven efficient for parameterized EM
design~\cite{gregory2011fast}, but its covariance model has
$\mathcal O(D^2)$ entries~\cite{hansen2016cma}, which rules it out at
the pixel level; the latent space brings $D$ into a practical range.
CMA-ES samples a population from a Gaussian, ranks the candidates by
their tuned losses $\widehat F$, and adapts the mean, covariance, and
step size from the ranking. Large latent regions decode to the same
topology, so the search is organized by layout families, i.e., groups
of masks with similar foreground. The first generation evaluates one
representative of each family in the dataset, and CMA-ES starts at the
best one. When progress stalls, the search
restarts~\cite{auger2005restart}, first at the incumbent with a smaller
step size and then in another family, keeping the best design found.

\subsection{Ensemble-Kalman Candidate Injection}
\label{subsec:general_kalman}

CMA-ES uses only a scalar ranking, although each evaluated design also
gives a vector of per-frequency specification residuals.
Ensemble-Kalman injection uses this vector to propose latent steps.
It adapts the ensemble Kalman
filter~\cite{evensen1994sequential} in its derivative-free form for
inverse problems~\cite{iglesias2013ensemble}, in which the gain is
estimated from ensemble covariances instead of Jacobians. Similar
updates have guided pretrained diffusion models with black-box forward
models~\cite{zheng2024ensemble}, where they move the generative samples
directly. Here, the Kalman step only proposes candidates, which CMA-ES
ranks together with its own samples.

For each candidate, the tuned response gives a residual vector
$\mathbf r$ with one signed, normalized entry per metric and
frequency, and a violation vector $\mathbf v(\mathbf r)$ that keeps
only the part of each entry outside its admissible range, so satisfied
specifications contribute no correction. From the search vectors and
residuals of the most recent generations, the ensemble cross-covariance
$\mathbf C_{xq}$ and residual covariance $\mathbf C_{qq}$ are
estimated, and a proposal is formed from an anchor $\mathbf x_{\mathrm a}$,
one of the best designs of the previous generation:
\begin{equation}
    \mathbf x_{\mathrm K}
    =
    \mathbf x_{\mathrm a}
    -
    \rho\,
    \mathbf C_{xq}
    (\mathbf C_{qq}+\mathbf R_{\mathrm a})^{-1}
    \mathbf v(\mathbf r_{\mathrm a}),
    \label{eq:general_kalman_gain}
\end{equation}
where $\rho\in(0,1)$ is a damping factor. The diagonal regularization
$\mathbf R_{\mathrm a}$ is smallest for the residual entries that
dominate the anchor's LSE loss, so the step targets the specifications
that currently limit it, and the step length is capped at that of a
typical CMA-ES sample. The proposals replace an equal number of
ordinary samples, so the EM budget per generation is unchanged, and
they enter the CMA-ES ranking only through their losses.

\section{Quadrature Coupler Design and Experimental Results}
\label{sec:experimental_results}

\begin{figure}[t!]
	\includegraphics[width=0.35\textwidth]{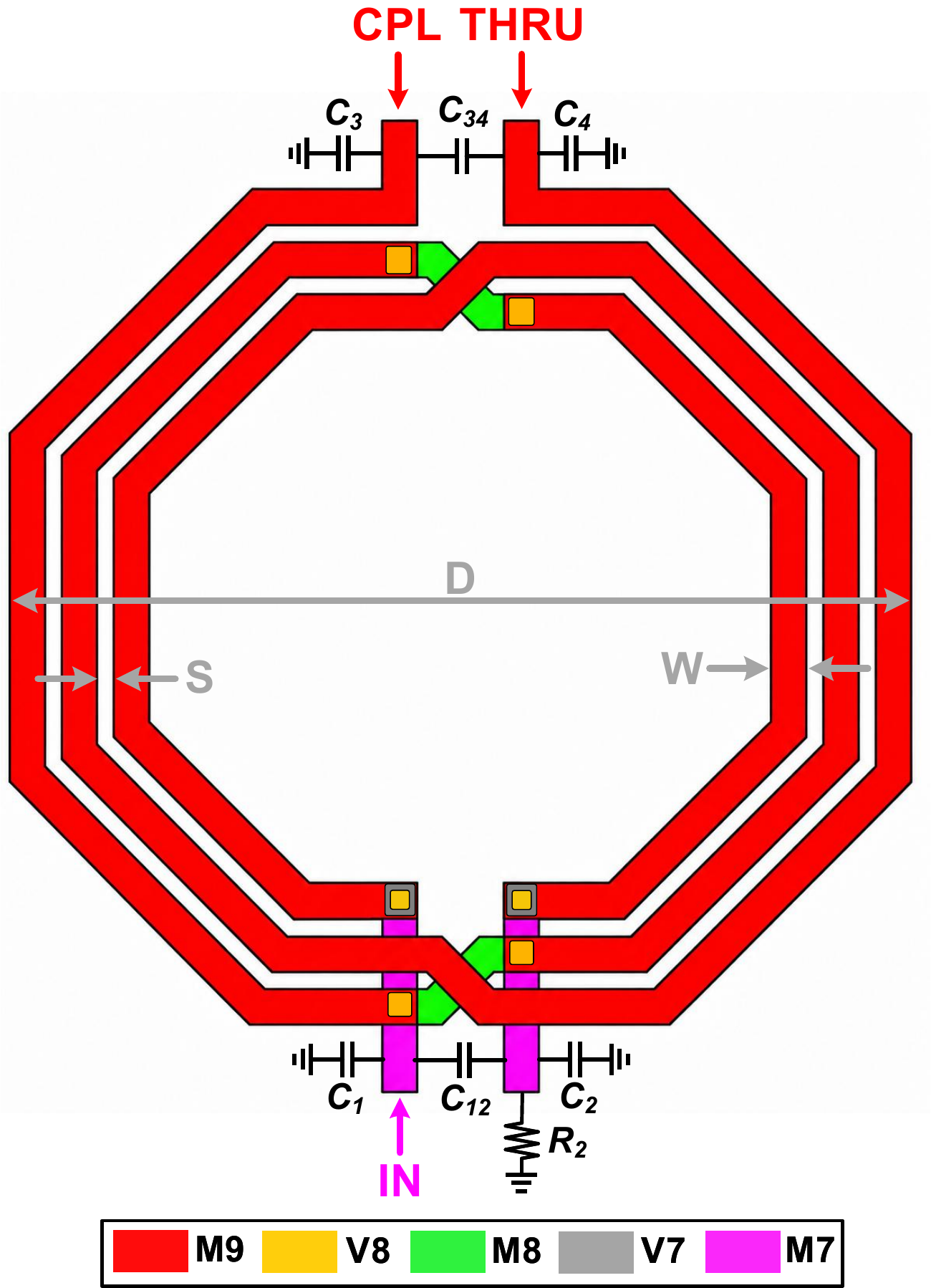}
	\centering
	\caption{Base layout of the QC, with the classic knobs: the
	outer diameter $D$, trace width $W$, and spacing $S$.}
	\label{fig:qc_base}
\end{figure}

The framework of Sections~\ref{sec:generation}--\ref{sec:optimization}
is applied to broadband QCs realized as
coupled-winding transformers (XFMRs). The reference
structure~\cite{Zolkov2026}, shown in Fig.~\ref{fig:qc_base}, consists
of two interleaved octagonal windings routed mainly on the top metal
(M9). The windings exchange radial position through two crossover
motifs, which use M8 and the V8 vias, and the input feed passes under
the windings on M7. One winding connects the input (IN) and isolated
ports, and the other connects the coupled (CPL) and through (THRU)
ports. The lumped elements in Fig.~\ref{fig:qc_base} are not part of
the drawn geometry but realize the port impedances of
Fig.~\ref{fig:LSO_diagram}: the shunt capacitors $C_1$--$C_4$ realize
$Z_1$--$Z_4$, the bridging capacitors $C_{12}$ and $C_{34}$ realize
$Z_{12}$ and $Z_{34}$, and the isolation resistor $R_2$ is added at
the isolated port. Their values form the circuit variables
$\boldsymbol\theta$ of
Section~\ref{subsec:general_electrical_objective}. The geometry of
this reference is described at three levels, following the preceding
sections: by the classic knobs of a textbook XFMR, by the local knobs
of Section~\ref{sec:generation}, and by a pixelated mask and its
learned latent representation
(Sections~\ref{sec:concept}--\ref{sec:optimization}).

\subsection{Classic Knobs}
\label{subsec:qc_classic}

A textbook parameterization of the XFMR
(Fig.~\ref{fig:algo_1st}) has three global parameters, marked in
Fig.~\ref{fig:qc_base}: the outer diameter $D$, the trace width $W$, and the
spacing $S$ between neighboring tracks. One width and one spacing
apply to every leg of both windings, both nets stay on M9 apart from
the M8 crossovers and the M7 feed, and there is no ground. On the
$63\times63$ grid of Section~\ref{subsec:qc_ablations}, with a
378-$\mu$m layout side, the three knobs are continuous within
$2\le W\le30~\mu$m, $2.5\le S\le40~\mu$m, and
$76\le D\le363~\mu$m. Every
choice of $(D,W,S)$ gives a valid layout, but the topology, the
uniform cross section, and the lateral coupling along the whole
winding are fixed.

\subsection{Local Knobs}
\label{subsec:qc_local}

\begin{figure*}[t!]
	\centering
	\begin{minipage}[t]{\columnwidth}
		\centering
		\includegraphics[width=\columnwidth]{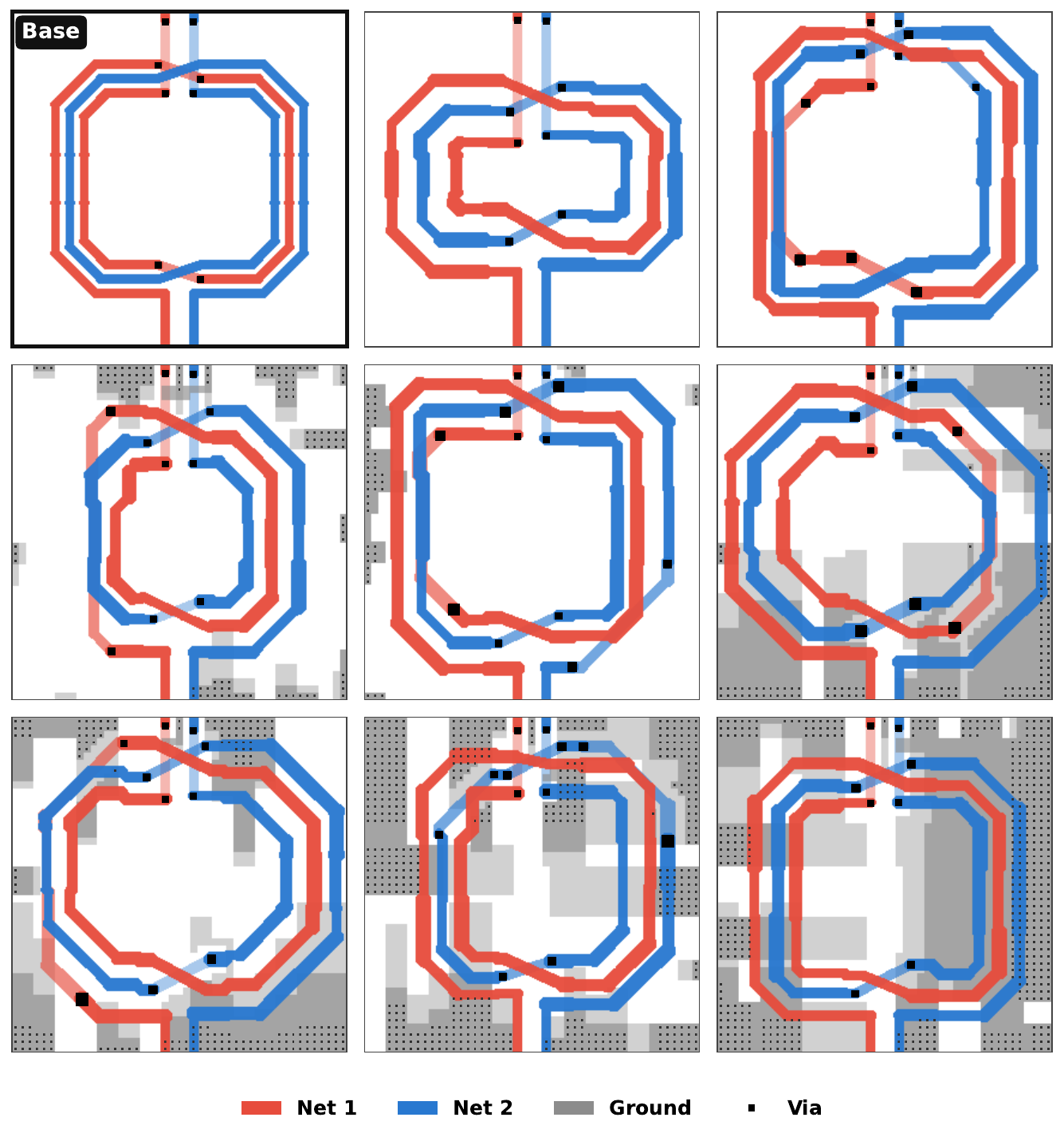}
		\caption{Reference layout of the QC (top left) and eight layouts
		produced by the local knobs in continuous coordinates.}
		\label{fig:qh47_param_knobs_examples}
	\end{minipage}\hfill
	\begin{minipage}[t]{\columnwidth}
		\centering
		\includegraphics[width=\columnwidth]{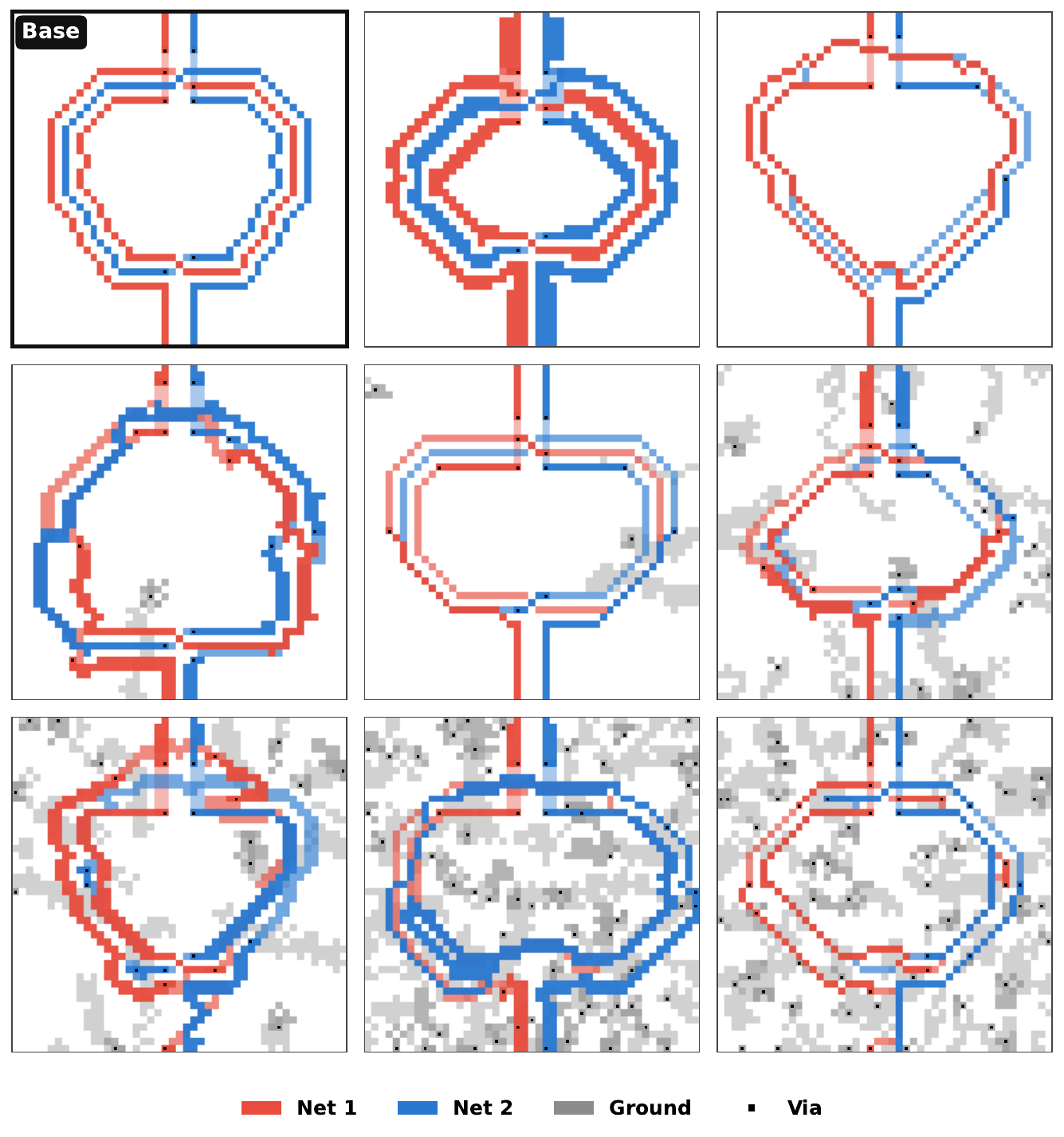}
		\caption{Pixelated reference mask of the QC on the $47\times47$ grid
		(top left) and eight masks produced by the mask generator.}
		\label{fig:qh47_v5_oldgnd_examples_3x3_seed1_top}
	\end{minipage}
\end{figure*}

The parameterization of Section~\ref{sec:generation} replaces these
three parameters with knobs on each part of the same reference. Each
net runs about one and a half turns around a chamfered octagon through
15 legs, divided into 27 sub-segments, each with its own width
(2--30~$\mu$m) and inset (up to a quarter of the layout side), so $W$
and $S$ vary along the winding. Net~2 mirrors the outline
and break positions of net~1 but has its own widths, insets, and
metals. Each leg is on M9, on M8 stacked under the other net, or on M8
at its own inset, so the coupling can change between lateral and
broadside from leg to leg; the crossover jogs and feeds can use M9,
M8, or M7. This gives 188 signal knobs: 150 continuous ones for the
outline, break positions, widths, insets, jogs, and feeds, and 38
discrete ones for the metal of each leg, jog, and feed.

The reference has no ground. The local knobs add it on M7 and M8,
below the windings, inward from the four edges of the layout area,
with V7 vias joining the two planes. Along each edge, the M7 depth is
a profile of a constant and 12 cosine harmonics, so the ground can
approach some parts of the windings and recede from others, and the
M8 ground and the vias reach set fractions of that depth. With an
on/off and a density knob, this gives 59 ground knobs, each taking one
of 2 to 11 discrete levels, for 247 knobs in all, 97 of them discrete.
The ground stays on the pixel grid and is trimmed to the minimum
spacing from the signal polygons, so it can load chosen parts of the
windings with capacitance without touching them
(Fig.~\ref{fig:generic_examples}(e)). Every knob vector is checked by
the rules of Section~\ref{subsec:general_validity}.
Fig.~\ref{fig:qh47_param_knobs_examples} shows the reference and eight
layouts produced by the local knobs.

\subsection{Pixelated Representation and Latent Space}
\label{subsec:qc_reference}

For the latent search, the same reference is pixelated on a
$47\times47$ or a $63\times63$ grid (Section~\ref{subsec:qc_implement})
with $L=5$ planes (M9, V8, M8, V7, and M7) and $K=4$ labels: empty
space, the two nets, and ground (top-left panel of
Fig.~\ref{fig:qh47_v5_oldgnd_examples_3x3_seed1_top}). The contract
fixes the four port pixels and reserves the boundary of M7 for ground,
and the validator $\mathcal V$ requires each net to connect its ports
as one component, isolated from the other net and from ground, every
via to land on metal on both sides, and every ground region to reach
the boundary of M7.

The mask generator produces the same kinds of layouts as the local
knobs but sets only a few of their features explicitly. Its structural
controls $\mathbf u$ are the vertical position of the two crossovers,
the lateral offset of the anchors on the sides of the winding, the
extensions of the crossovers, a width class, and a layer plan that
puts each net on M9 or M8 at five stations, as lateral, broadside, or
mixed coupling. What the local knobs set per sub-segment comes instead
from random choices: the windings are routed between the anchors by
the A$^*$ search of Section~\ref{subsec:general_pixel_generator} and
widened by random growth, so their width and position vary from
section to section, and the ground is placed by the edge profiles of
Section~\ref{subsec:qc_local} with random parameters. The other panels
of Fig.~\ref{fig:qh47_v5_oldgnd_examples_3x3_seed1_top} range from
near-reference layouts to strongly deformed windings with wider
traces, broadside sections, and more ground.

The validated masks train the AE and the structural predictor
(Appendix~\ref{app:ae}), separately for each grid. The latent grid is
$12\times12$ for the $47\times47$ masks and $16\times16$ for the
$63\times63$ masks, giving 144 or 256 search variables. The predictor
estimates the controls $\mathbf u$ except the layer plan, which is read
from the logits.

\begin{figure}[t!]
	\centering
	\includegraphics[width=\columnwidth]{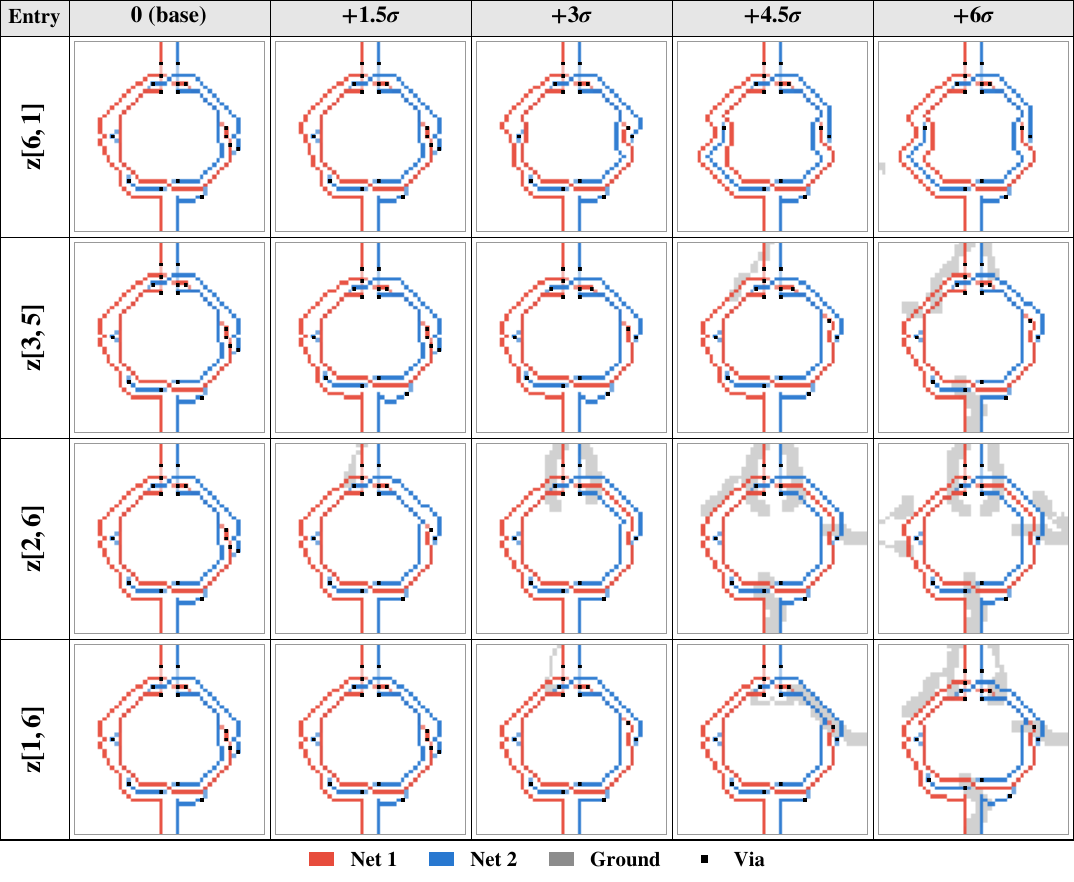}
	\caption{Latent traversal of the $47\times47$ QC. Each row raises one
	entry $z[y,x]$ of the $12\times12$ latent from 0 to $+6\sigma$ in steps
	of $1.5\sigma$, starting from the same base latent $\mathbf z_0$; all
	other entries are fixed. Top view of all metal layers stacked.}
	\label{fig:qh47_latent_traversal}
\end{figure}

Fig.~\ref{fig:qh47_latent_traversal} shows how the $47\times47$ layout
responds to single entries of its $12\times12$ latent $\mathbf z$,
which encodes both the signal routing and the ground. Every row starts
from the same base latent $\mathbf z_0$, the encoding of a ground-free
training layout close to the latent mean. Each row increases one entry
$z[y,x]$ from 0 to $+6\sigma$ in steps of $1.5\sigma$, where $\sigma$
is the standard deviation of that entry over the training set, while
all other entries stay fixed. Each panel is a top view of all metal
layers stacked, with Net~1 in red, Net~2 in blue, ground in gray, and
vias as black squares. Moving $z[6,1]$ mainly reshapes the winding,
pulling its left side inward. Moving $z[3,5]$ first shifts the
crossings and then adds ground near the feed lines. Moving $z[2,6]$ and
$z[1,6]$ grows ground steadily, first next to the feeds and then along
the sides of the winding, while the routing changes only slightly. The
geometry changes gradually along each row, and each entry controls a
recognizable, mostly local feature of the layout.

The local knobs and the latent space describe overlapping families of multilayer geometries, with approximately the same number of optimization variables: 247 knobs versus 256 latent entries for the $63\times63$ grid (144 for the $47\times47$ grid). However, the two representations are not identical. The local knobs define continuous polygonal geometries with explicit geometric parameters, whereas the latent representation describes pixelated geometries generated through learned routing and structural variations. Consequently, differences in optimization performance may arise from both the structure of the search spaces and the geometries accessible through each representation. The knobs are a mixture of 150 continuous and 97 discrete
variables, each tied to one feature, such as the width of one
sub-segment or the metal of one leg. Features that act together, such
as the insets of two facing tracks, which set their gap, or the widths
of neighboring sub-segments, are separate knobs that the optimizer
must move together, and a discrete knob changes the layout in a jump.
The latent entries are all continuous and are arranged on a grid that
follows the layout, so each one moves a local group of correlated
features, such as a side of the winding, a crossing, or a ground
region, gradually and consistently
(Fig.~\ref{fig:qh47_latent_traversal}). The two spaces also differ in
what they can express. The knobs draw exact polygons on a fixed route
of legs, sub-segments, and $45^\circ$ jogs, with sub-pixel widths
(Fig.~\ref{fig:qh47_param_knobs_examples}). The
latent layouts lie on the pixel grid, but their windings are routed
freely between the anchors, so their width and position can change
from pixel to pixel
(Fig.~\ref{fig:qh47_v5_oldgnd_examples_3x3_seed1_top}), and the AE
learns from the training layouts which of these variations occur
together.

\subsection{Specifications and Optimization Setup}\label{subsec:qc_implement}
Three XFMR QCs were designed with LSO
(Section~\ref{sec:optimization}) for the 10--20, 25--50, and
30--60~GHz bands and fabricated in 65-nm CMOS. The 10--20~GHz design
uses the $63\times63$ grid with $6\times6~\mu\mathrm{m}^2$ pixels, and
the 25--50 and 30--60~GHz designs use the $47\times47$ grid with
$5\times5~\mu\mathrm{m}^2$ and $4\times4~\mu\mathrm{m}^2$ pixels,
respectively. The underlying quadrature coupler is a four-port network, with its isolated port terminated on chip, leaving three externally accessible ports. Measurements are performed using a two-port vector network analyzer (VNA). A shunt 50-$\Omega$
termination is therefore added on chip at each port, so any pair of
ports can be probed while the third remains terminated (Section~\ref{subsec:qc_measurements}).
All capacitors are
metal-oxide-metal (MOM) structures. In all three bands, the key
performance indicators (KPIs) are matching below $-10$~dB at all
ports, phase imbalance below $2^\circ$, and amplitude imbalance below
1~dB, with the insertion loss (IL) minimized.

The KPIs are translated into the loss of
Section~\ref{subsec:general_electrical_objective} as follows. With
the IN port excited, let $S_{\mathrm T}(f_k)$ and $S_{\mathrm C}(f_k)$
denote the transmissions to the THRU and CPL ports at the $k$th
simulated frequency $f_k$, with magnitudes
$T_k=20\log_{10}|S_{\mathrm T}(f_k)|$ and
$C_k=20\log_{10}|S_{\mathrm C}(f_k)|$ in dB, and let $S_{pp}(f_k)$
denote the reflection coefficient of port $p$. The per-frequency
metrics are
\begin{equation}
    \begin{aligned}
    a_k &= \left|T_k-C_k\right|,\\
    \varphi_k &= \min_{\pm}\big|
        \operatorname{wrap}\!\left(\angle S_{\mathrm T}(f_k)-\angle S_{\mathrm C}(f_k)
        \mp 90^\circ\right)\big|,\\
    l^{\mathrm T}_k &= \left|T_k+3\right|,\qquad
    l^{\mathrm C}_k = \left|C_k+3\right|,\\
    \gamma^{p}_k &= \max\!\left(0,\;20\log_{10}|S_{pp}(f_k)|-\Gamma_{\mathrm{th}}\right),
    \end{aligned}
    \label{eq:qc_metrics}
\end{equation}
for amplitude imbalance, phase error from quadrature, deviation of
each output from the ideal $-3$~dB split, and reflection at port $p$,
respectively, where $\Gamma_{\mathrm{th}}$ is the matching threshold.
Here $\angle$ denotes the phase angle, the phase error is taken to the
nearer of $\pm90^\circ$, and $\operatorname{wrap}(\cdot)$ maps angles to
$(-180^\circ,180^\circ]$. Because a
passive coupler cannot exceed $-3$~dB at both outputs, driving
$l^{\mathrm T}_k$ and $l^{\mathrm C}_k$ to zero minimizes the IL.

Let bold symbols collect a metric over all frequencies, e.g.,
$\mathbf a=[a_1,a_2,\ldots]$ and $\boldsymbol\gamma^p=[\gamma^p_1,\gamma^p_2,\ldots]$.
The metrics are aggregated over frequency with
(\ref{eq:general_softmax_loss}) and normalized by their tolerances
$A_0$, $P_0$, $L_0$, and $R_0$, giving the category terms
\begin{equation}
    \begin{aligned}
    E_A &= \operatorname{LSE}_{\alpha}(\mathbf a)/A_0,\qquad
    E_P = \operatorname{LSE}_{\alpha}(\boldsymbol\varphi)/P_0,\\
    E_L &= \operatorname{LSE}_{\alpha}\!\left(
        \operatorname{LSE}_{\alpha}(\mathbf l^{\mathrm T}),\,
        \operatorname{LSE}_{\alpha}(\mathbf l^{\mathrm C})\right)/L_0,\\
    E_R &= \operatorname{LSE}_{\alpha}\!\left(
        \{\operatorname{LSE}_{\alpha}(\boldsymbol\gamma^{p})\}_{p\in\mathcal P_R}
        \right)/R_0,
    \end{aligned}
    \label{eq:qc_categories}
\end{equation}
for amplitude imbalance, phase, output magnitude, and reflection, where
$\mathcal P_R$ is the set of ports whose reflection is scored. They are
combined into the scalar loss
\begin{equation}
    \mathcal L_{\mathrm{QC}}
    =
    \operatorname{LSE}_{\beta}\!\left(E_A,E_P,E_L,E_R\right),
    \label{eq:qc_loss}
\end{equation}
where $\beta$ is the sharpness across categories. $\mathcal L_{\mathrm{QC}}$
is the loss $\mathcal L$ of (\ref{eq:general_bilevel}) for the QC.
The tolerances $A_0=1$~dB and $P_0=2^\circ$ equal the imbalance
KPIs, so $E_A$ and $E_P$ reach one at the specification limit; we
also use $L_0=1.25$~dB and $R_0=1$~dB. The matching threshold
$\Gamma_{\mathrm{th}}=-11$~dB leaves a 1-dB margin to the matching KPI. The reflection set PR includes the THRU and CPL ports, while input matching is not explicitly included in the optimization objective. Input matching is evaluated separately in the full-wave simulation and verified through the measured results. The sharpness parameters $\alpha=20$ and $\beta=5$ make
(\ref{eq:qc_loss}) a smooth approximation of the worst frequency in
the worst category.

Each candidate mask is extracted once with an EM solver. Two pixels of
the same net that touch only at a corner would form a single-point
contact, which violates design rules and is ill defined for the
solver, so a diamond-shaped patch is drawn over each such corner
during layout conversion~\cite{chenna2026inverse}. NM then
tunes $\boldsymbol\theta$ through circuit simulation, minimizing
(\ref{eq:qc_loss}) as in (\ref{eq:general_bilevel}). For the
ensemble-Kalman proposals, the residual vector
$\mathbf r$ concatenates the signed amplitude
imbalance, phase error, and output-magnitude deviations at every
frequency as exact targets, normalized by $A_0$, $P_0$, and $L_0$.
It also includes the signed reflection margins
$(20\log_{10}|S_{pp}|-\Gamma_{\mathrm{th}})/R_0$ as one-sided
requirements, so only reflections above the threshold contribute
to the violation $\mathbf v$.

The micrographs are shown in
Figs.~\ref{fig:meas_10to20GHz}(a), \ref{fig:meas_25to50GHz}(a), and
\ref{fig:meas_30to60GHz}(a); the active areas are 0.16, 0.067, and
0.044~mm$^2$. Each QC was optimized in no more than 2~hours on a
64-core AMD Ryzen Threadripper PRO 7985WX workstation; the AE and the structural
predictor ran on a single NVIDIA RTX A400 GPU.

\subsection{Measurements}\label{subsec:qc_measurements}
The fabricated chips were measured on-wafer using a VNA and two
100-$\mu$m-pitch ground-signal-ground (GSG) probes. The 50-$\Omega$
on-chip terminations were de-embedded from the measured S-parameters.
For all three chips, the measurements follow the simulated trends
with no visible frequency shift; within the design bands, the measured
IL is higher by about 0.2--0.3~dB.

\begin{figure}[t!]
	\centering
	\subfloat[]{{\includegraphics[width=0.23\textwidth]{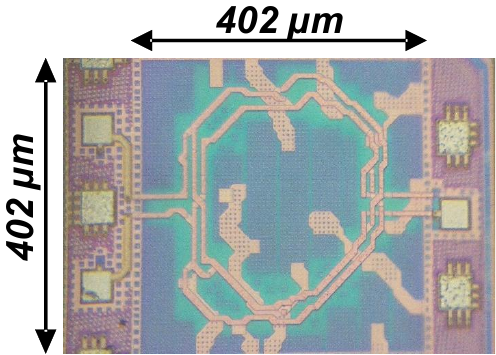}\label{fig:10to20GHz_TO}}}
	\quad
	\subfloat[]{{\includegraphics[width=0.23\textwidth]{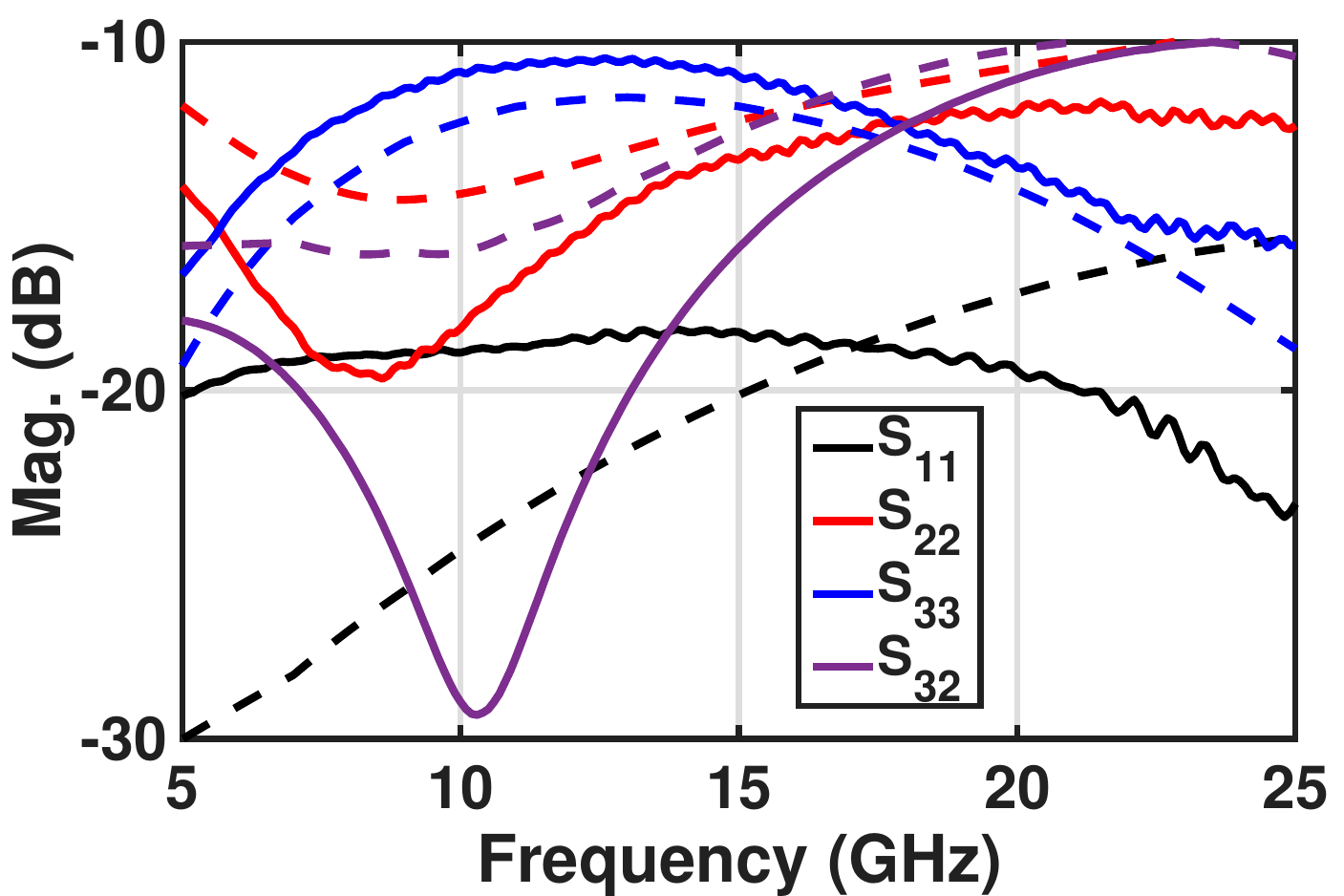}\label{fig:matching_10to20GHz}}} \\
	\centering
	\subfloat[]{{\includegraphics[width=0.25\textwidth]{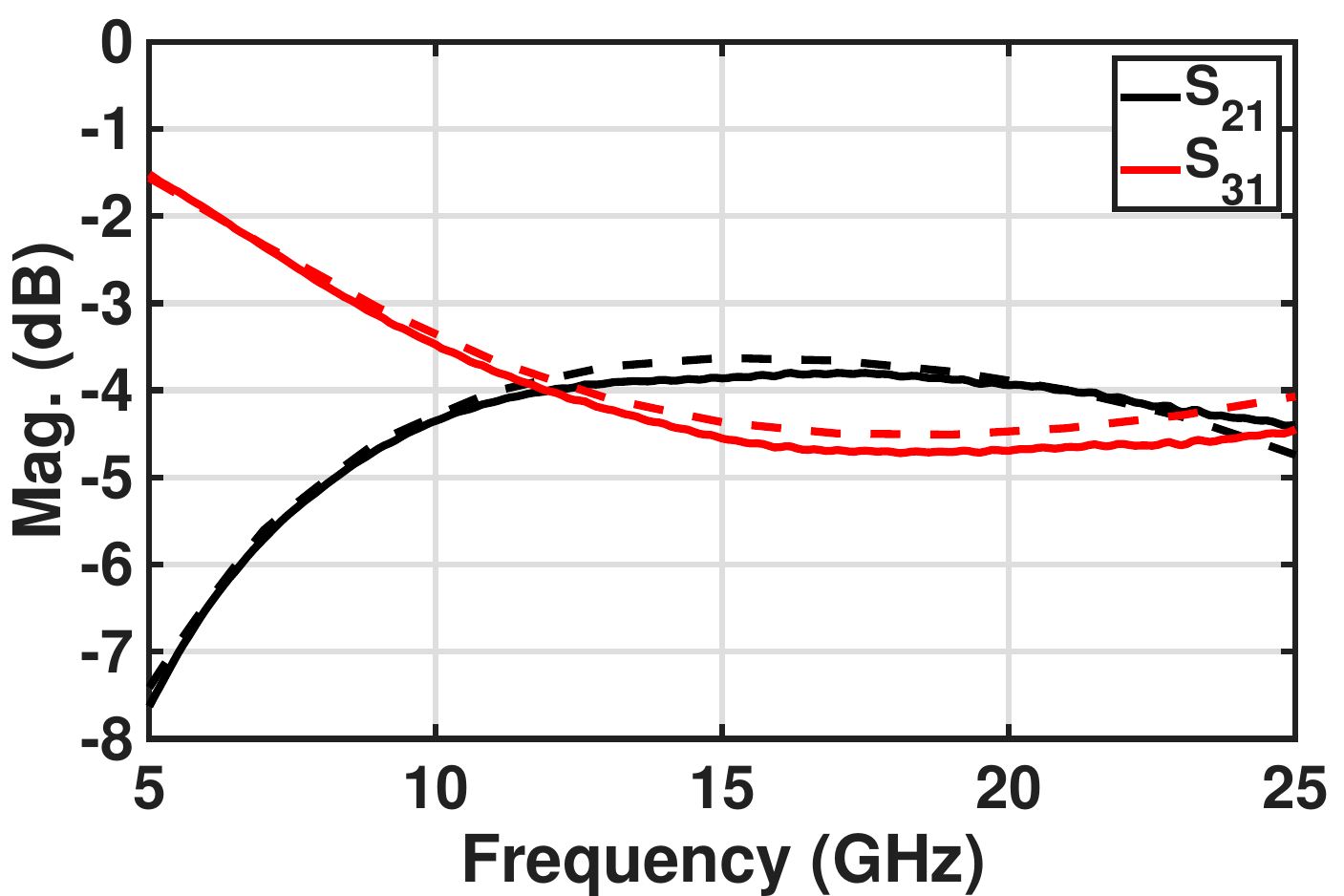}\label{fig:s21_s31_10to20GHz }}}
	\subfloat[]{{\includegraphics[width=0.25\textwidth]{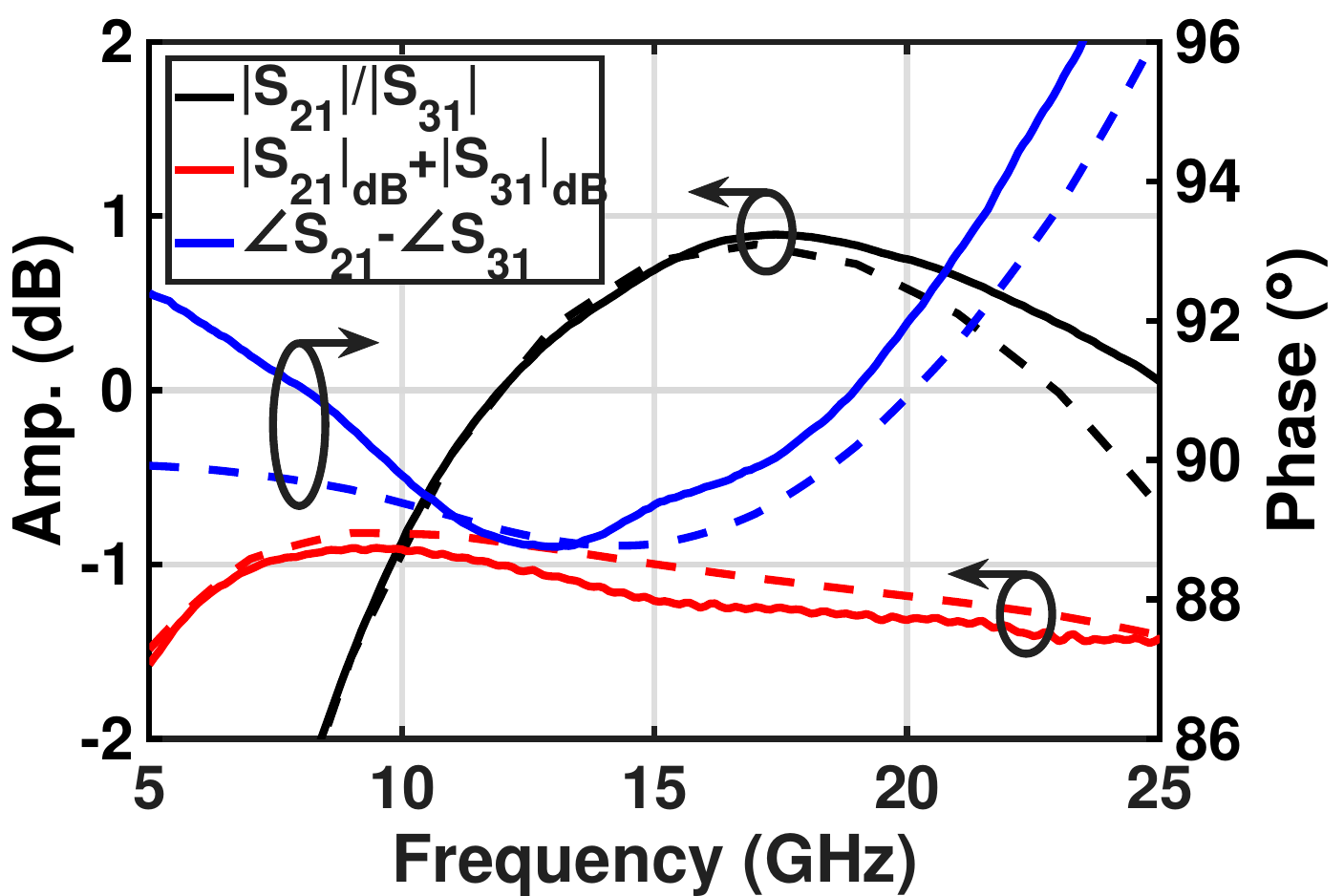}\label{fig:imbalance_il_10to20GHz}}}
	\caption{10 to 20 GHz QC (a) micrograph, and simulation (dashed) and measurement (solid) results of: (b) matching, (c) forward gain, and (d) amplitude imbalance, phase imbalance and IL.}\label{fig:meas_10to20GHz}
\end{figure}

Fig.~\ref{fig:meas_10to20GHz} shows the results of the 10--20~GHz
chip. Over the band, the measured reflection is below $-10$~dB at all
ports, and the chip has less than 1.4-dB IL, less than 1-dB amplitude
imbalance, less than 2$^\circ$ phase imbalance, and more than 11-dB
isolation between the output ports.

\begin{figure}[t!]
	\centering
	\subfloat[]{{\includegraphics[width=0.23\textwidth]{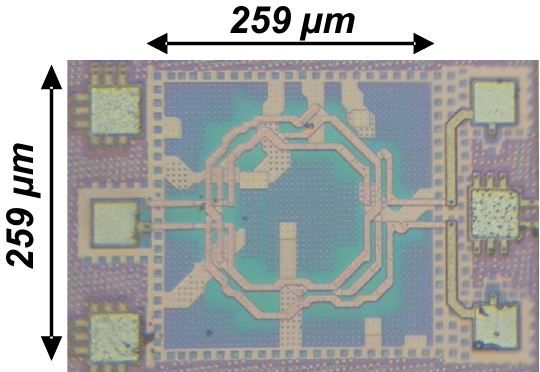}\label{fig:25to50GHz_TO}}}
	\quad
	\subfloat[]{{\includegraphics[width=0.23\textwidth]{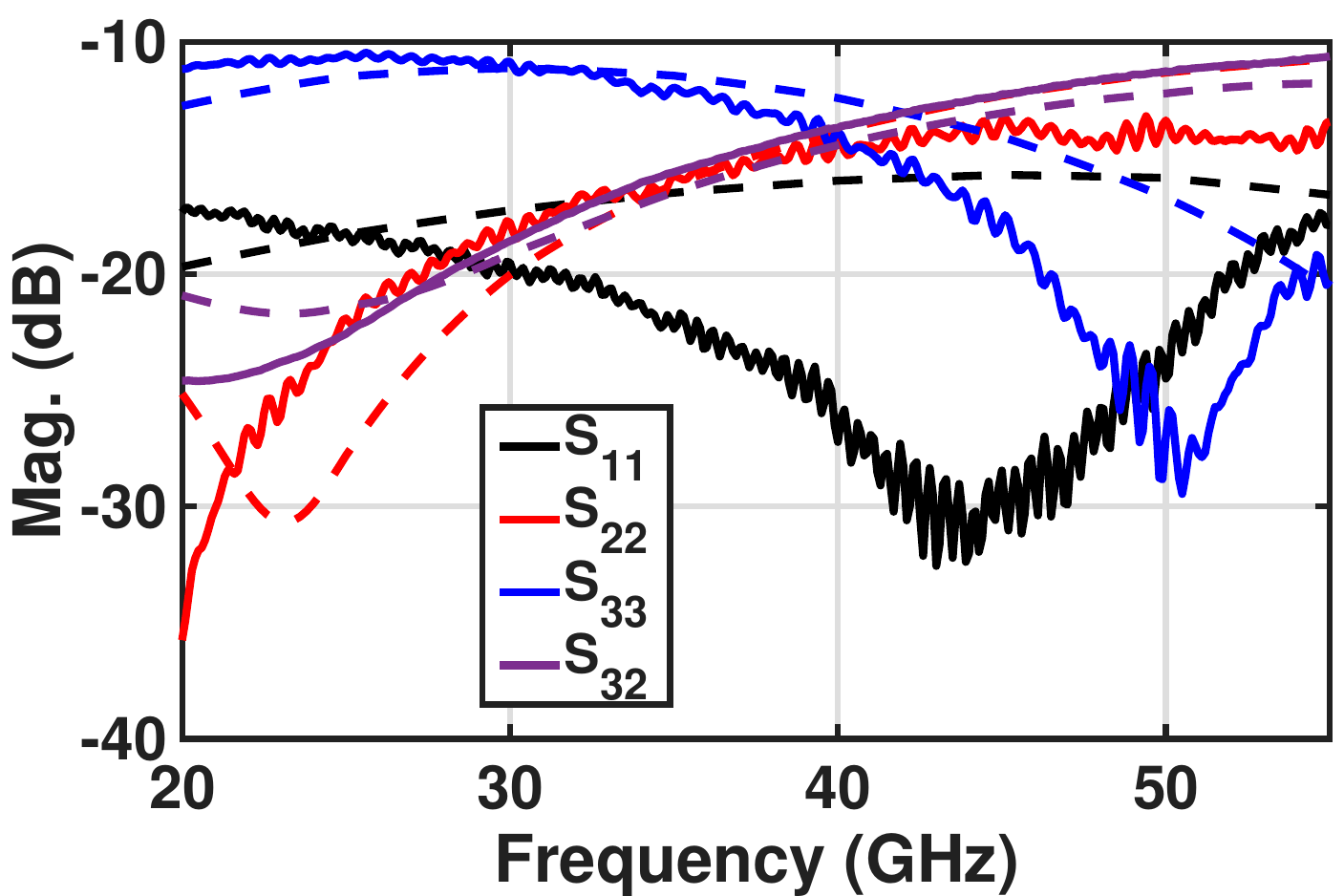}\label{fig:matching_25to50GHz}}} \\
	\centering
	\subfloat[]{{\includegraphics[width=0.25\textwidth]{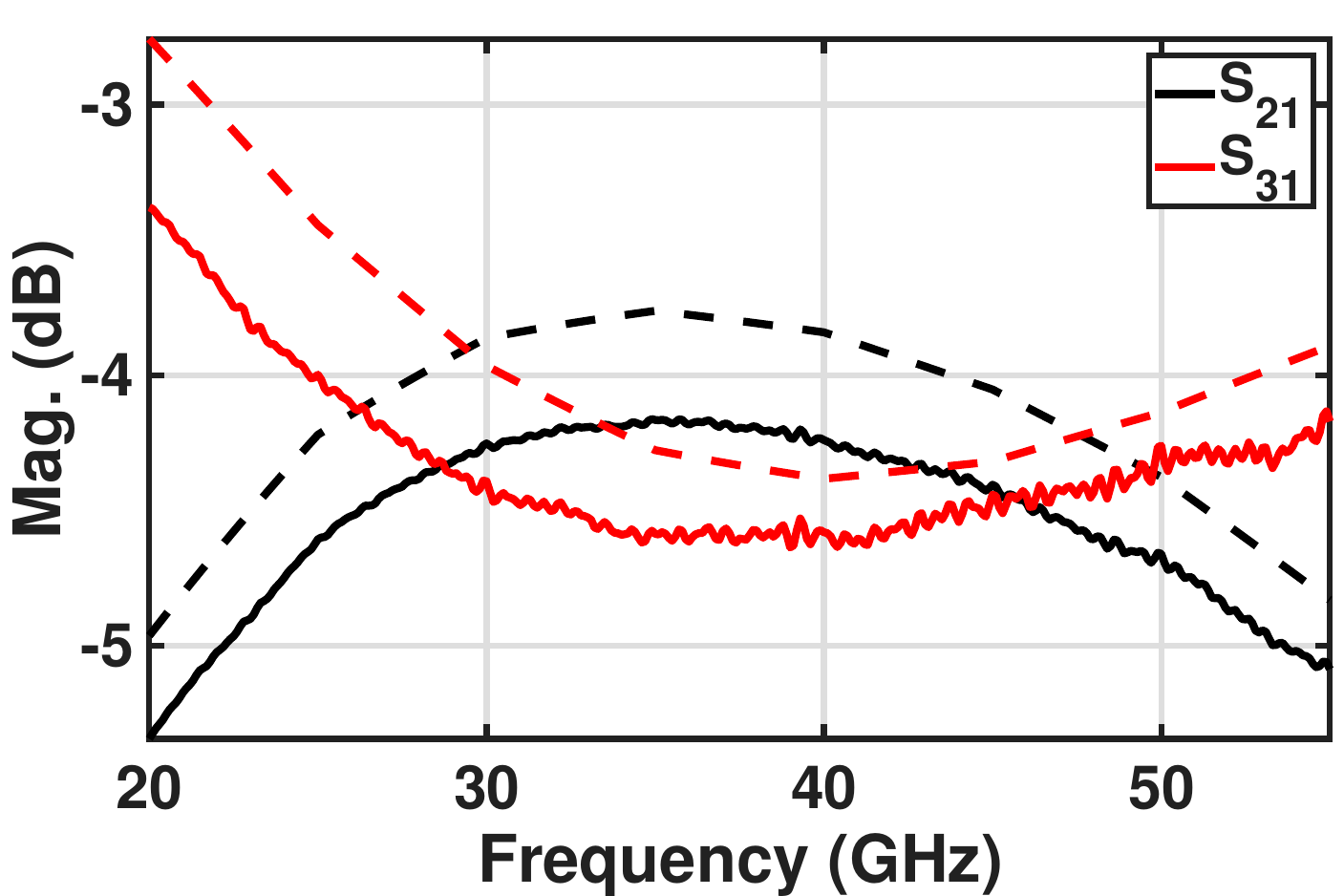}\label{fig:s21_s31_25to50GHz }}}
	\subfloat[]{{\includegraphics[width=0.25\textwidth]{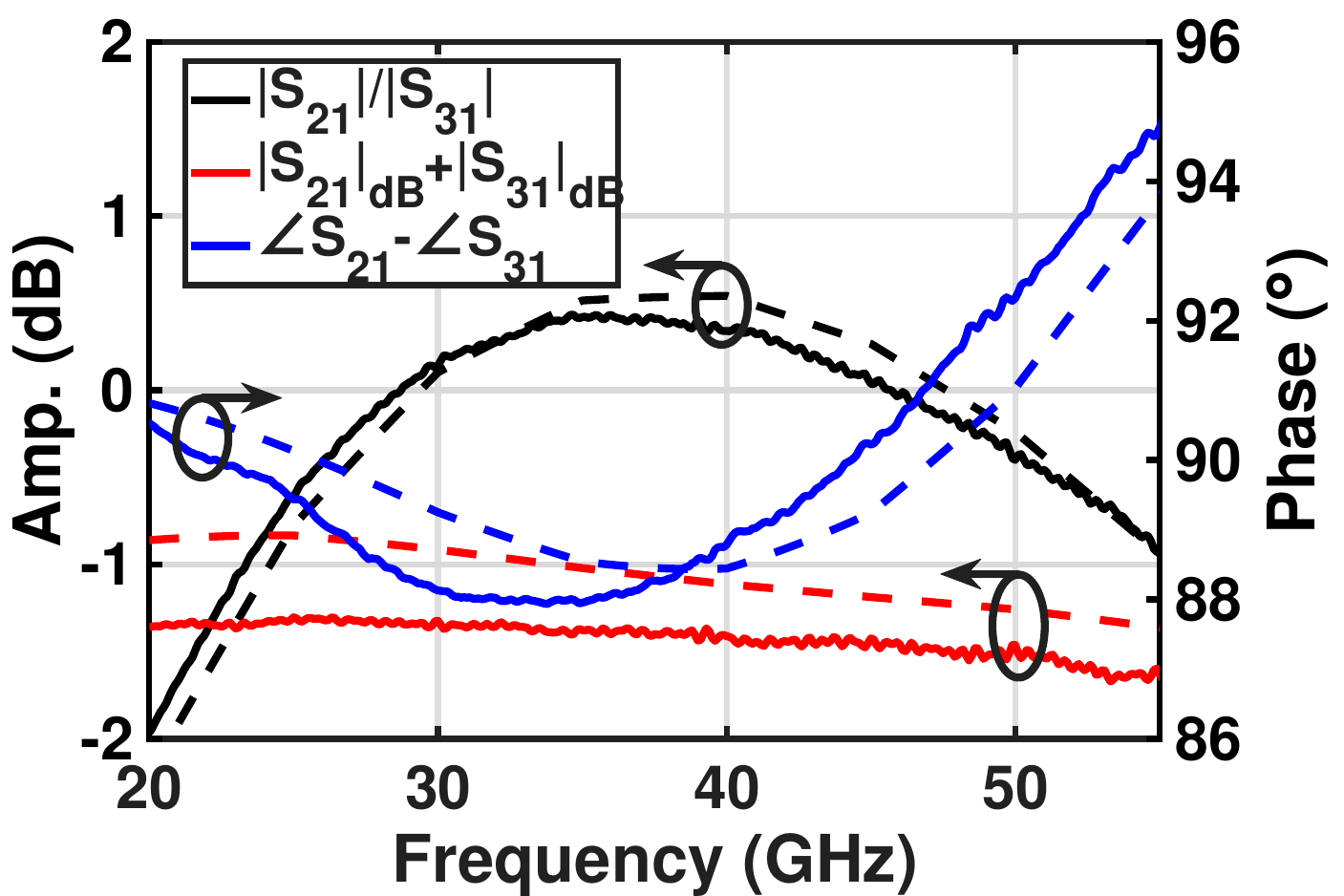}\label{fig:imbalance_il_25to50GHz}}}
	\caption{25 to 50 GHz QC (a) micrograph, and simulation (dashed) and measurement (solid) results of: (b) matching, (c) forward gain, and (d) amplitude imbalance, phase imbalance and IL.}\label{fig:meas_25to50GHz}
\end{figure}

Fig.~\ref{fig:meas_25to50GHz} shows the results of the 25--50~GHz
chip. The measured reflection is below $-10$~dB at all ports, and the
chip has less than 1.5-dB IL, less than 0.8-dB amplitude imbalance,
less than 2$^\circ$ phase imbalance, and more than 11-dB isolation.

\begin{figure}[t!]
	\centering
	\subfloat[]{{\includegraphics[width=0.23\textwidth]{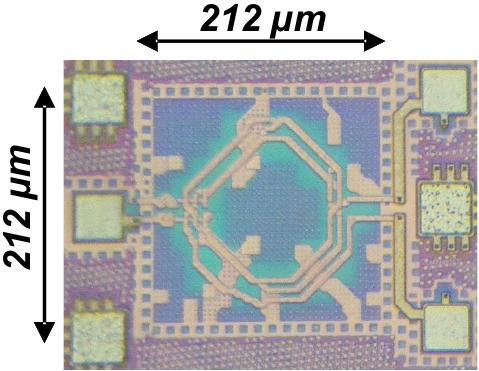}\label{fig:30to60GHz_TO}}}
	\quad
	\subfloat[]{{\includegraphics[width=0.23\textwidth]{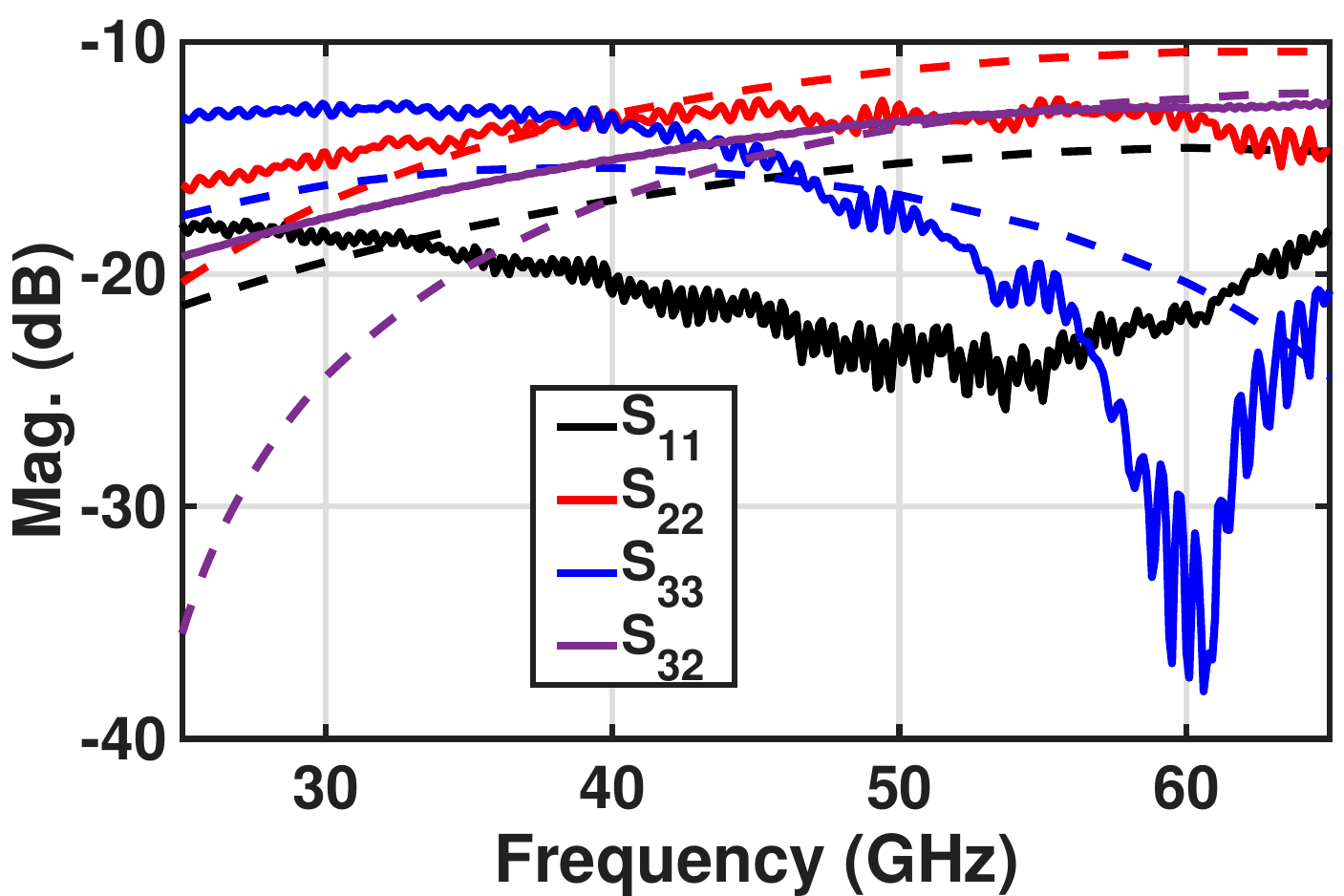}\label{fig:matching_30to60GHz}}} \\
	\centering
	\subfloat[]{{\includegraphics[width=0.25\textwidth]{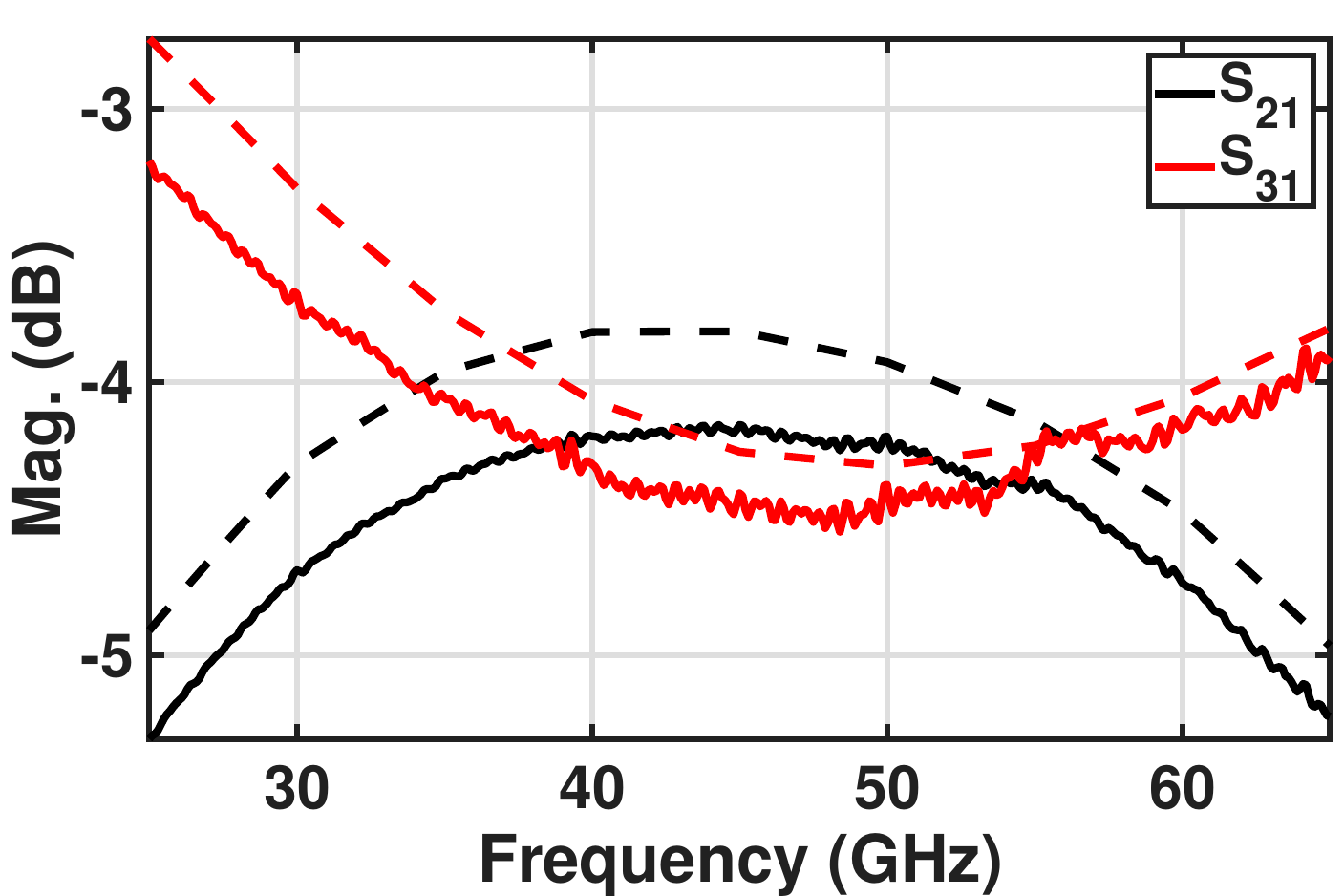}\label{fig:s21_s31_30to60GHz }}}
	\subfloat[]{{\includegraphics[width=0.25\textwidth]{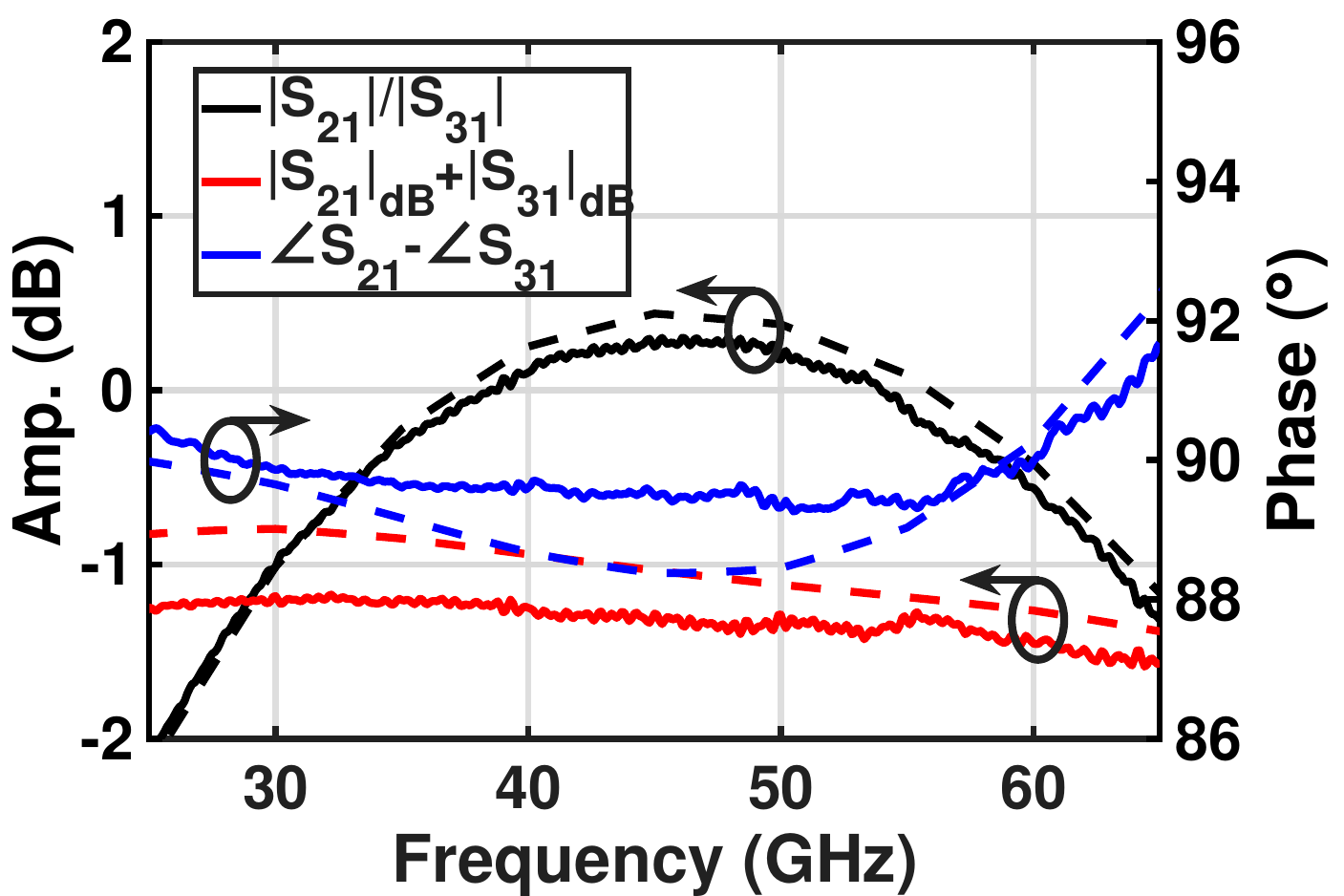}\label{fig:imbalance_il_30to60GHz}}}
	\caption{30 to 60 GHz QC (a) micrograph, and simulation (dashed) and measurement (solid) results of: (b) matching, (c) forward gain, and (d) amplitude imbalance, phase imbalance and IL.}\label{fig:meas_30to60GHz}
\end{figure}

Fig.~\ref{fig:meas_30to60GHz} shows the results of the 30--60~GHz
chip. The measured reflection is below $-10$~dB at all ports, and the
chip has less than 1.45-dB IL, less than 1-dB amplitude imbalance,
less than 1$^\circ$ phase imbalance, and more than 13-dB isolation.

\subsection{Comparison With Prior Art}
\label{subsec:qc_comparison}

Table~\ref{tab:comp_table} compares the proposed chips with
previously reported cascaded rat-race~\cite{Garay},
slow-wave coupler~\cite{Parveg}, slow-wave branch-line~\cite{Acri}, and
all-pass-filter~\cite{Qiu} implementations. All three proposed
chips cover a 67\% fractional bandwidth. The
25--50 and 30--60 GHz chips occupy 0.067 and 0.044~mm$^2$,
respectively, compared with 0.63, 0.16, 0.066, and 0.027~mm$^2$ for
the four references, while maintaining IL below 1.5 dB and phase
imbalance no greater than 2$^\circ$. The slow-wave branch-line coupler
of~\cite{Acri} occupies a similar area to the 25--50 GHz chip but has
a $<$2.35-dB IL and a narrower fractional
bandwidth.
The 10--20 GHz chip matches the 1.4-dB loss of the slow-wave coupler
reference in the same 65-nm CMOS process, and its phase imbalance is below
2$^\circ$, compared with 5$^\circ$ for the cascaded rat-race reference.  The
measured isolation of the proposed chips ($>$11--13 dB) is lower than
the 15--20 dB reported in~\cite{Garay,Parveg,Acri}, since isolation
was not included in the loss of (\ref{eq:qc_loss}); it can be added
as a further category term.

\definecolor{MyGreen}{RGB}{0,210,0}
\definecolor{Gray}{gray}{0.85}
\newcolumntype{P}[1]{>{\centering\arraybackslash}m{#1}}
\newcolumntype{F}[1]{>{\raggedleft}m{#1}}
\newcolumntype{G}[1]{>{\centering\arraybackslash\columncolor{Gray}}m{#1}}
\begin{table*}[t!]
\centering
\caption{Comparison With Prior Integrated QC Works}
\scriptsize
\centering
\begin{tabular}{!{\vrule width 1.25pt}G{0.15\textwidth}!{\vrule width 0.7pt}P{0.1\textwidth}!{\vrule width 0.7pt}P{0.1\textwidth}!{\vrule width 0.7pt}P{0.1\textwidth}!{\vrule width 0.7pt}P{0.1\textwidth}
!{\vrule width 0.7pt}P{0.06\textwidth}!{\vrule width 0.7pt}P{0.06\textwidth}!{\vrule width 0.7pt}P{0.06\textwidth}
!{\vrule width 1.25pt}
}
\noalign{\hrule height 1.25pt}
\rowcolor{Gray}
\textbf{Features} & \textbf{IMS 18' \cite{Garay}} & \textbf{MWCL 19' \cite{Parveg}} & \textbf{SSCL 19' \cite{Acri}} & \textbf{TCAS II 21' \cite{Qiu}} & \multicolumn{3}{c!{\vrule width 1.25pt}}{\textbf{This work}}\\
\noalign{\hrule height 0.7pt}
\textbf{Architecture} & Cascaded rat-race & Slow-wave coupler & Slow-wave branch-line & All-pass filter & \multicolumn{3}{c!{\vrule width 1.25pt}}{Inverse designed XFMR w/ caps}\\
\noalign{\hrule height 0.7pt}
\textbf{Process} & 45 nm CMOS SOI & 65 nm CMOS & 22 nm CMOS FDSOI & 65 nm CMOS & \multicolumn{3}{c!{\vrule width 1.25pt}}{65 nm CMOS}\\
\noalign{\hrule height 0.7pt}
\textbf{BW (GHz)} & 22-44 (67\%)& 55-110 (67\%)& 51-68 (29\%) & 20-40 (67\%) & 10-20 (67\%)& 25-50 (67\%)& 30-60 (67\%)\\
\noalign{\hrule height 0.7pt}
\textbf{IL (dB)} & $<$2.5 & $<$1.4 & $<$2.35 & $<$4.5 & $<$1.4 & $<$1.5 & $<$1.45\\
\noalign{\hrule height 0.7pt}
\textbf{Amplitude imbalance (dB)} & $<$1 & $<$1 & 0.8 & $<$0.3 & $<$1 & $<$0.8 & $<$1\\
\noalign{\hrule height 0.7pt}
\textbf{Phase imbalance ($^\circ$)} & $<$5 & $<$2.5 & 1.8 & $<$2 & $<$2 & $<$2 & $<$1\\
\noalign{\hrule height 0.7pt}
\textbf{Isolation (dB)} & $>$15 & $>$20 & $>$16 & N/R & $>$11 & $>$11 & $>$13\\
\noalign{\hrule height 0.7pt}
\textbf{Active area (mm$\mathit{\mathbf{^{2}}}$)} & 0.63 & 0.16 & 0.066 & 0.027 & 0.16 & 0.067 & 0.044 \\
\noalign{\hrule height 1.25pt}
\end{tabular}
\\
\vspace{2pt}
\centering
\scriptsize{N/R: Not reported}
\label{tab:comp_table}
\end{table*}

\subsection{Optimization Efficiency and Comparison}
\label{subsec:qc_ablations}

\begin{figure}[t!]
	\centering
	\subfloat[]{{\includegraphics[width=0.25\textwidth]{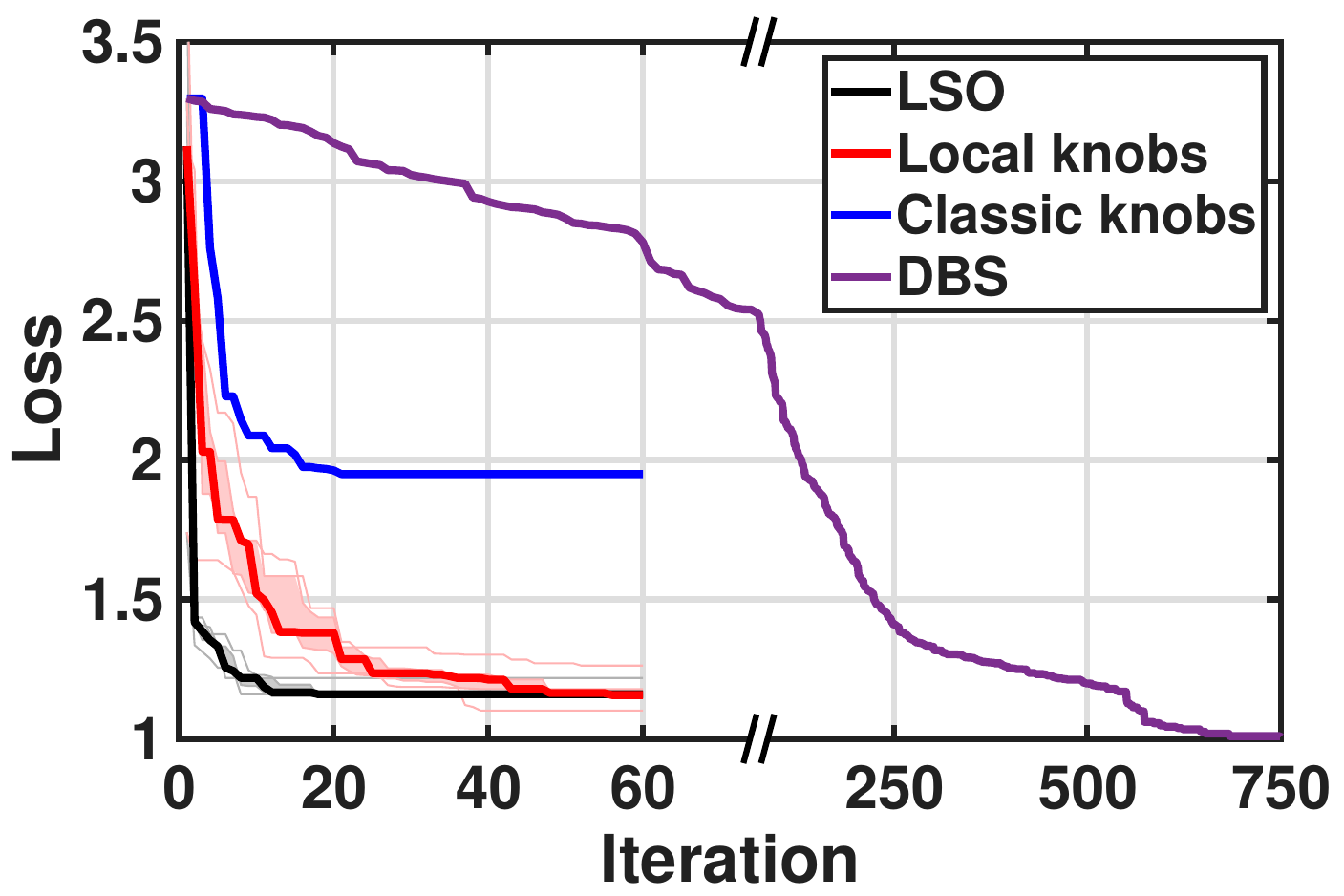}\label{fig:opt_comp_loss_vs_gen}}}
	\subfloat[]{{\includegraphics[width=0.25\textwidth]{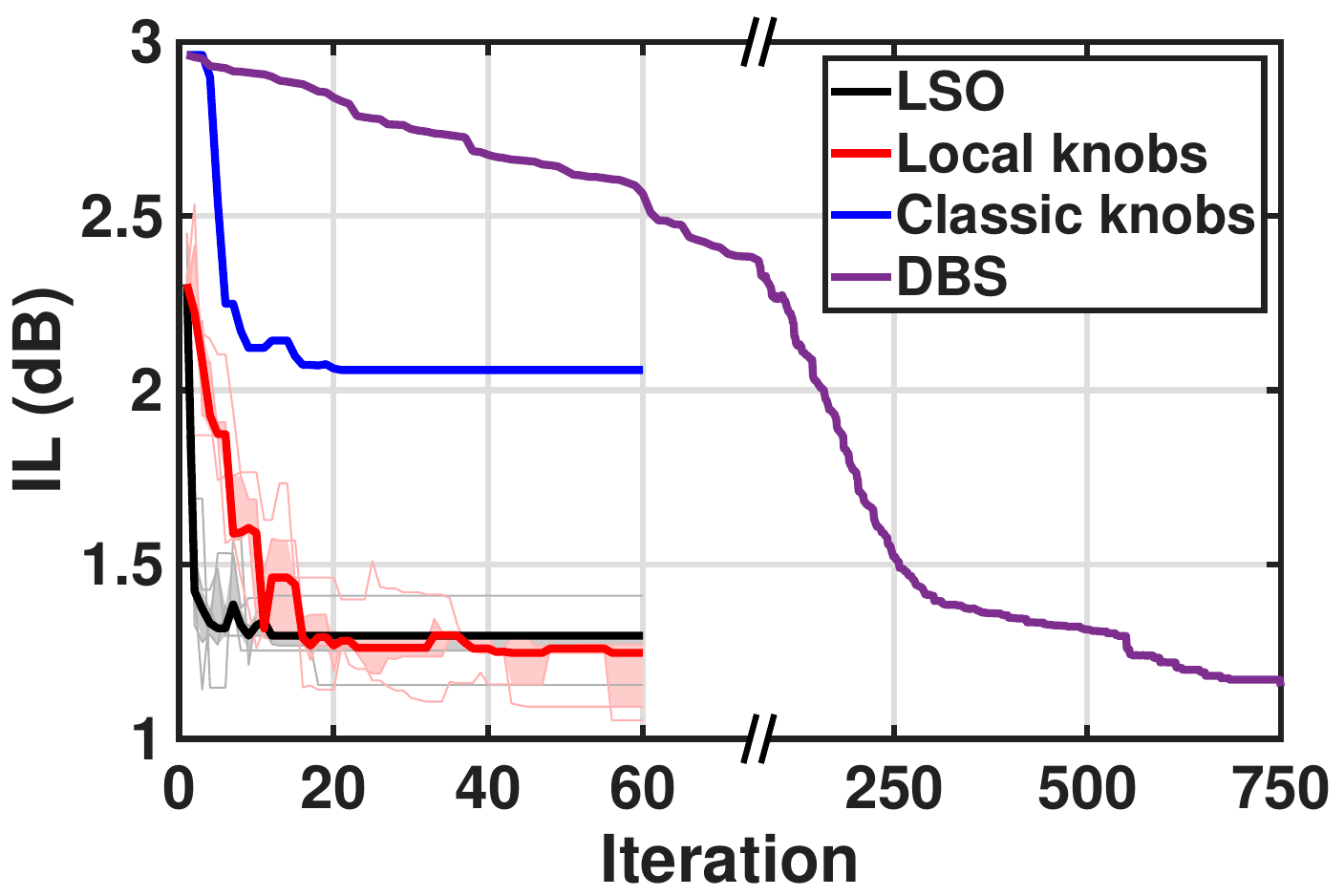}\label{fig:opt_comp_il_vs_gen}}} \\
	\centering
	\subfloat[]{{\includegraphics[width=0.25\textwidth]{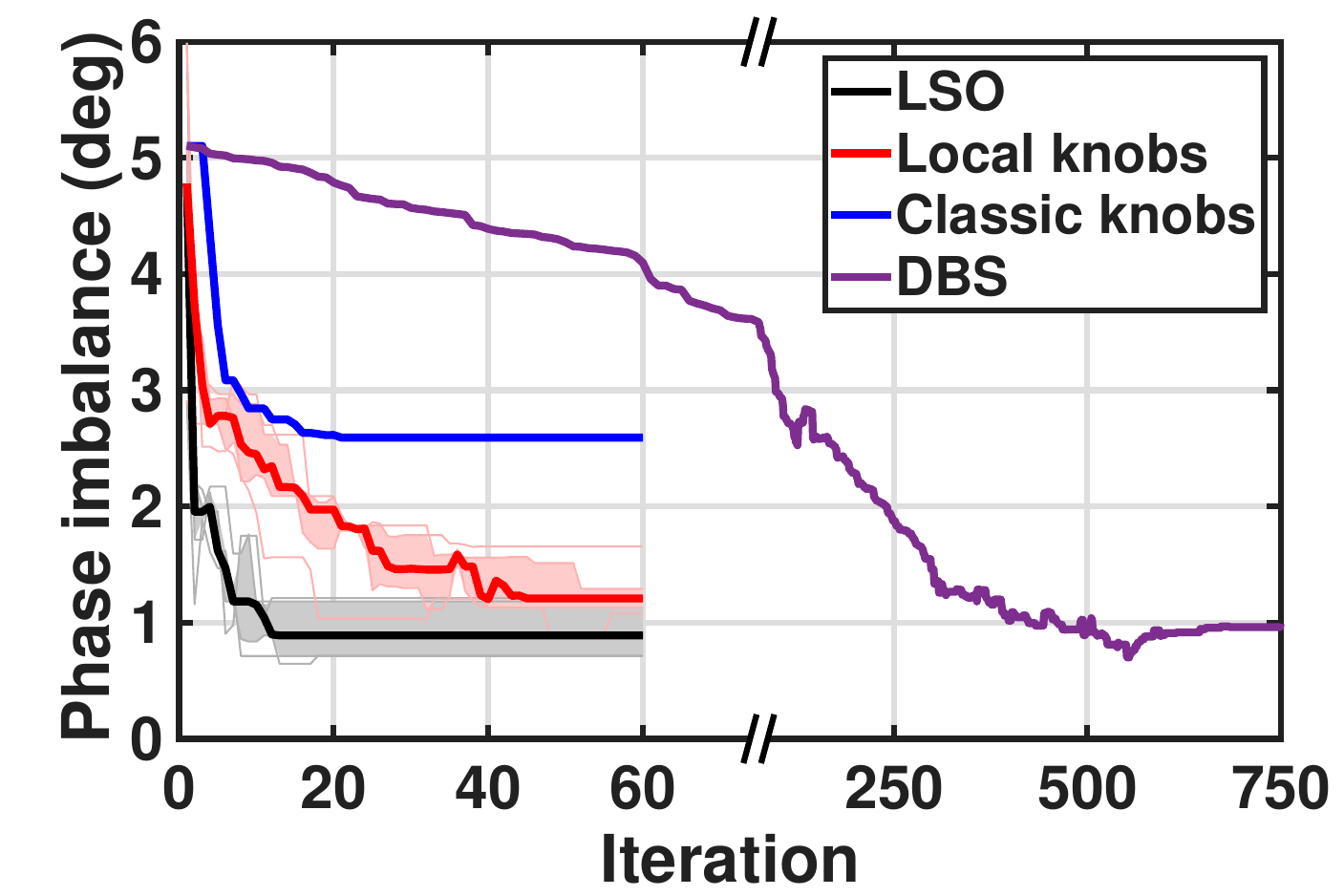}\label{fig:opt_comp_phase_imb_vs_gen}}}
	\subfloat[]{{\includegraphics[width=0.25\textwidth]{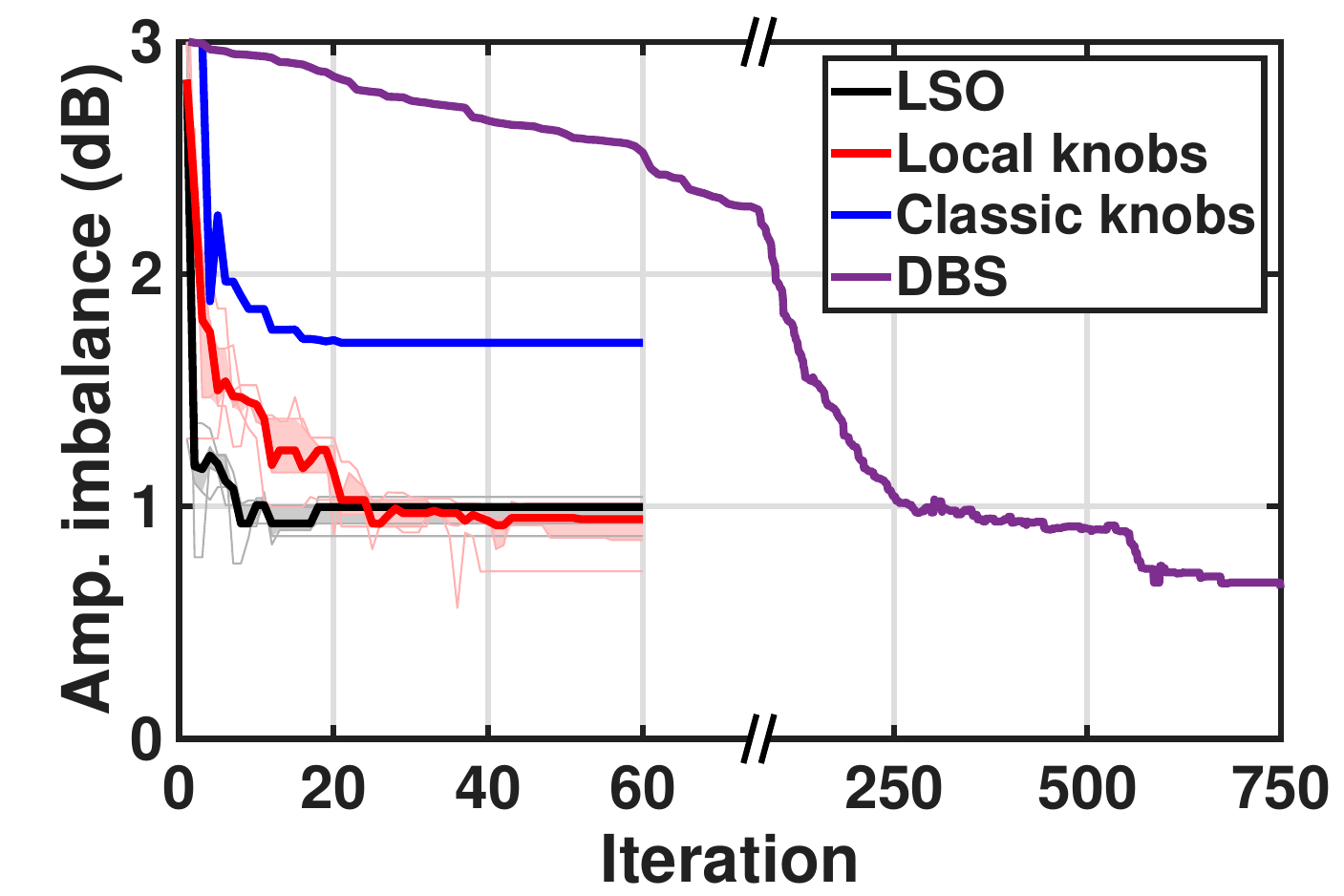}\label{fig:opt_comp_amp_imb_vs_gen}}}
	\caption{Optimization runs of the 10--20~GHz QC: DBS
	of the pixels, CMA-ES on the classic knobs ($D$, $W$, and $S$ of
	Fig.~\ref{fig:qc_base}), CMA-ES on the local knobs, and LSO, each from
	20 seed layouts with 20 EM simulations per iteration. LSO and the local
	knobs are each run five times: thin lines show the individual runs,
	thick lines the median, and the shaded band the interquartile range.
	Best design so far:
	(a) loss, and worst case over the band of (b) IL, (c) phase
	imbalance, and (d) amplitude imbalance. The iteration axis is
	compressed twentyfold beyond iteration 65 (break mark); only DBS runs
	past iteration 60, up to 750.}\label{fig:opt_comparison}
\end{figure}

To separate the two steps of the framework, the 10--20~GHz QC
($63\times63$ grid, $6\times6~\mu\mathrm{m}^2$ pixels) was optimized
with four searches: DBS of the pixels,
CMA-ES on the classic knobs of the XFMR, CMA-ES on the local knobs of
Section~\ref{sec:generation}, and LSO. Every run uses the same EM
solver, NM circuit tuning, and loss (\ref{eq:qc_loss}), evaluates
20 candidates per iteration, and starts from 20 seed layouts,
built with both nets on the top metal and no ground. LSO and the local
knobs were each run five times, with different CMA-ES random seeds and
from five sets of seed layouts shared by the two searches; the DBS and classic-knob runs start from one of
these sets. The two knob searches
and LSO use the same CMA-ES with ensemble-Kalman injection
(Section~\ref{sec:optimization}) and restarts, and differ only in the
vector being searched. These runs
were stopped after 60 iterations (1200 EM simulations); the DBS
run, which improves far more slowly, was continued to 750 (15000). An
iteration takes about 5~minutes, so LSO, whose median loss reaches 1.2
at iteration 11, needs about 55~minutes, within the 2-hour budget of the
fabricated designs. Fig.~\ref{fig:opt_comparison} shows the best
design found so far in each run. For LSO and the local knobs, the
losses and KPIs given below are medians over the five runs.

\textit{1) Classic knobs.} This run tunes the three classic knobs $D$,
$W$, and $S$ of Section~\ref{subsec:qc_classic}
(Fig.~\ref{fig:qc_base}). The run stalls at a loss of 1.95 after iteration 21, with 2.1-dB IL,
1.7-dB amplitude imbalance, and 2.6$^\circ$ phase imbalance, missing
both imbalance KPIs: with the topology and a uniform cross section
fixed, three parameters leave too little freedom over an octave.

\textit{2) Local knobs.} This run tunes the 247 knobs of
Section~\ref{subsec:qc_local}. Both knob runs use the polygons,
validity checks, and EM export of Section~\ref{sec:generation}, so
they differ only in their knobs. In both, infeasible vectors are
redrawn, discrete knobs are rounded, and CMA-ES starts from the
parametric fit of the best seed. The local-knob search passes the final classic loss at
iteration 5, reaches 1.2 at iteration 43, and ends at 1.16 at
iteration 56, meeting both imbalance KPIs with 1.25-dB IL.

\textit{3) DBS.} In this run, DBS~\cite{chenna2026pixelated} keeps one current
layout, initialized with the best seed. In each iteration, 20 mutants
are formed by flipping one to ten random non-port pixels and checked
by the validator, and the best replaces the current layout if it
lowers the loss. After 1200 EM simulations the loss has dropped only
from 3.3 to 2.8, with IL and amplitude imbalance near 2.6~dB and phase
imbalance above 4$^\circ$. The run passes the final classic loss after
2700 simulations and a loss of 1.2 after 9960. The local-knob search
reaches this loss after 860 simulations, 12 times fewer, and LSO after
220, 45 times fewer. Most pixel flips in
a multilayer winding either break a net or barely change the response,
whereas a single knob changes a whole feature of the layout: the
metal of a leg, the width of a sub-segment, or a ground patch. After
all 15000 simulations, about 62~hours, DBS ends at 1.01, with 1.15-dB
IL, 0.65-dB amplitude imbalance, and 0.99$^\circ$ phase imbalance.

\textit{4) LSO.} This run searches the 256-entry latent vector, and each
candidate is rebuilt by the guided generator of
Section~\ref{subsec:general_guidance}. It passes the final classic loss
at iteration 2 and reaches 1.2 after 220 simulations (iteration 11),
3.9 times fewer than the 860 of the local knobs. It then levels off
at 1.16 from iteration 18, the final loss that the local knobs reach
only at iteration 56. LSO ends with the lower phase imbalance,
0.89$^\circ$ versus 1.21$^\circ$, and the local knobs with the lower
worst-case IL, 1.25 versus 1.30~dB, and amplitude imbalance, 0.95
versus 1.01~dB, the latter at the 1-dB limit. Since the two searches have approximately the same number of variables (256 and 247), dimensionality reduction alone cannot explain the observed improvement. The results are consistent with the continuous, spatially correlated structure of the learned latent space, in which changes to individual variables produce coordinated geometric variations (Fig.~\ref{fig:qh47_latent_traversal}). In contrast, the direct parameterization combines continuous and discrete knobs whose interactions must be learned during optimization. Differences between the geometry families accessible through the two representations may also contribute to the observed convergence behavior.

\section{Conclusion}\label{sec:conclusion}
This article presented a two-step framework for the inverse design of
multilayer EM structures that bridges the global parameters of
textbook topologies and the pixels of free-form inverse design. 

In the first step, a constrained generator with local knobs allows the metal layer, conductor width, and position to vary from section to section, while also introducing ground patches. Geometric validation and regeneration ensure that the resulting layouts satisfy the prescribed connectivity and geometric constraints. In the
second, an AE trained on generated layouts, without EM simulation,
replaces the knobs and pixels with a continuous latent space, far
smaller than the pixel grid.
Each latent candidate is rebuilt by the generator, guided by the
decoder output and a learned predictor of the structural controls, and
is evaluated by full-wave EM simulation with jointly tuned lumped
circuit elements. CMA-ES searches the latent space, and
ensemble-Kalman proposals use the per-frequency specification
residuals that a scalar ranking discards. The optimization framework is not specific to quadrature couplers; extending it to other passive EM structures requires adapting the geometry generator, validity constraints, and electrical objective to the target application.

The framework was demonstrated with three XFMR-based QCs fabricated
in 65-nm CMOS, each optimized in no more than 2~hours. The designs
cover the 10--20, 25--50, and 30--60~GHz bands with measured ILs below
1.5~dB, amplitude imbalances below 1~dB, phase imbalances below
$2^\circ$, and reflections below $-10$~dB, in active areas of 0.16,
0.067, and 0.044~mm$^2$. In optimization comparisons using shared initial layouts, LSO reached a loss of 1.2 after a median of 220 EM simulations across five runs, compared with 860 simulations for direct local-knob optimization and 9,960 simulations for the representative DBS run. Thus, LSO required approximately 3.9 times fewer EM simulations than local-knob optimization and 45 times fewer than DBS to reach the same objective value. LSO also reached a loss of 1.16 after 18 iterations, whereas direct local-knob optimization reached this value after 56 iterations. The latent and local-knob representations have similar dimensionalities, suggesting that the improved convergence is associated with the continuous, spatially correlated structure of the learned representation rather than dimensionality reduction alone, although differences in the accessible geometries may also contribute. The search on the
local knobs, in turn, reached a far lower loss than a search on the
classic width, spacing, and diameter, which stalled short of the
imbalance KPIs, and needed about 12 times fewer EM simulations than
DBS to reach the same specified loss. Each of the two steps, the local
parameterization and the learned representation, thus improves the
search.

\appendices
\section{AE and Structural Predictor Details}
\label{app:ae}

\subsection{Training Objective}
\label{app:ae_objective}

Signal pixels are reconstructed exactly, with a per-pixel cross-entropy
$\mathcal L_{\mathrm{CE}}(\boldsymbol\pi,c)=-\log\pi_c$ for predicted
class probabilities $\boldsymbol\pi$ and target class $c$. Ground is
supervised only as a soft occupancy field through a term
$\mathcal L_{\mathrm{gnd}}$ on its local density and total mass.
Three regularizers prepare the latent space for optimization. First,
Gaussian noise is added to the latent code during training,
\begin{equation}
    \mathbf z'
    =
    \mathbf z
    +
    \eta_\tau s\boldsymbol\epsilon,
    \qquad
    \boldsymbol\epsilon
    \sim\mathcal N(\mathbf0,\mathbf I),
    \label{eq:general_noise}
\end{equation}
where $s$ is the standard deviation of the latent minibatch and the
noise ratio $\eta_\tau$ is ramped up during training. Denoising
AEs~\cite{vincent2008extracting} corrupt the input; here the code is
corrupted, since the optimizer moves in that space. A variational
AE~\cite{kingma2014auto} was not used, since its prior trades
reconstruction accuracy for sampling, which the search does not need.
Second, a
differentiable flood-fill penalty $\mathcal L_{\mathrm{conn}}$
discourages disconnected signal regions and inconsistent vias in the
decoded probabilities. Third, smoothness and cycle terms keep the
decoded output, and its re-encoding, stable under small latent
perturbations. The complete objective is
\begin{equation}
    \mathcal L_{\mathrm{AE}}
    =
    \mathcal L_{\mathrm{CE}}^{\mathrm{sig}}
    +\lambda_{\mathrm{g}}\mathcal L_{\mathrm{gnd}}
    +\lambda_{\mathrm{conn}}\mathcal L_{\mathrm{conn}}
    +\lambda_{\mathrm{reg}}\mathcal L_{\mathrm{reg}},
    \label{eq:general_ae_objective}
\end{equation}
where $\mathcal L_{\mathrm{CE}}^{\mathrm{sig}}$ is the signal
cross-entropy averaged over all pixels, $\mathcal L_{\mathrm{reg}}$
collects the noise, smoothness, and cycle terms, and the $\lambda$
coefficients weight the terms.

\subsection{Network Architecture}
\label{app:ae_architecture}

\begin{figure}[t!]
	\includegraphics[width=\columnwidth]{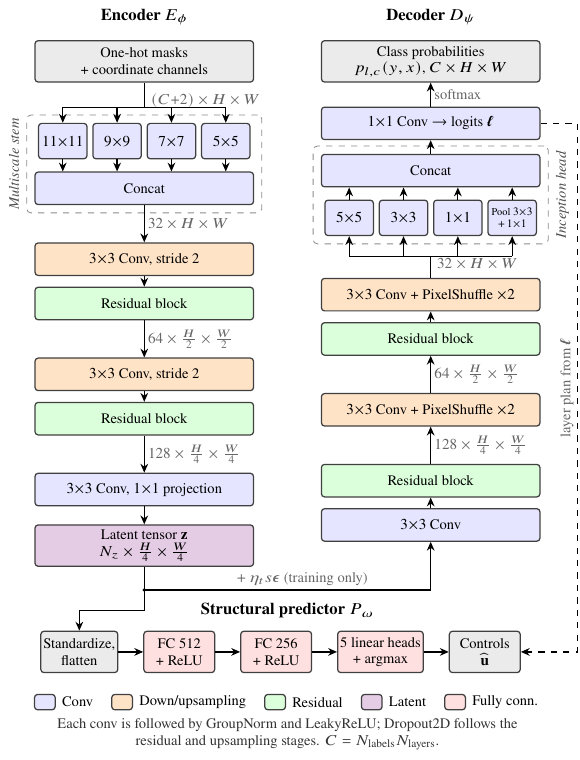}
	\centering
	\caption{AE architecture and structural predictor. The encoder (left) maps the one-hot
	mask to a single latent tensor $\mathbf z$ at one quarter of the
	grid resolution; the decoder (right) expands it and returns per-pixel
	class probabilities. The predictor $P_\omega$ (bottom) maps the
	standardized latent vector to the structural controls
	$\widehat{\mathbf u}$; the layer plan is read from the logits
	$\boldsymbol\ell$ (dashed). Shapes are channels $\times$ height $\times$ width.}
	\label{fig:ae_architecture}
\end{figure}

The AE (Fig.~\ref{fig:ae_architecture}) is a convolutional network
with a multiscale input stem~\cite{szegedy2015going}, residual
blocks~\cite{he2016deep}, group normalization~\cite{wu2018group},
coordinate channels~\cite{liu2018intriguing}, and sub-pixel
upsampling~\cite{shi2016real}. For the QC, the input contains
$LK=20$ one-hot channels and two coordinate channels. The stem's
$11\times11$ to $5\times5$ kernels capture features from single traces
to whole winding sections. Two stride-2 stages then
reduce the grid by a factor of four, and a $3\times3$ convolution
followed by a $1\times1$ projection forms a single latent tensor
$\mathbf z$ with $N_z=1$ channel. The decoder applies a
$3\times3$ convolution to $\mathbf z$, restores the full grid
through two PixelShuffle stages, and refines the result in an
inception head that returns the logits $\boldsymbol\ell$. The AE
is trained with the objective of (\ref{eq:general_ae_objective})
and is then frozen.

\subsection{Structural Predictor}
\label{app:predictor}

The predictor $P_\omega$ is trained on a dataset
$\mathcal D_u=\{\widetilde{\mathbf z}^{(n)},\mathbf u^{(n)}\}_{n=1}^{N_u}$
of masks generated with forced controls and encoded by the frozen AE,
by minimizing
\begin{equation}
    \mathcal L_{\mathrm{struct}}
    =
    \sum_{j=1}^{J}
    \lambda_j
    \mathcal L_{\mathrm{CE}}
    (\widehat{\boldsymbol\pi}_j,u_j),
    \label{eq:general_knob_loss}
\end{equation}
where $J$ is the number of predicted controls, $u_j$ is the target
class of control $j$, $\widehat{\boldsymbol\pi}_j$ is its predicted
class distribution, and $\lambda_j$ is its weight. For the QC, the
predictor, shown at the bottom of Fig.~\ref{fig:ae_architecture}, is
a multilayer perceptron with two fully connected (FC) hidden layers of
512 and 256 units and $J=5$ classification heads: the crossover
position, the side-anchor offset, the two crossover extensions, and
the width class. The width is a class because the generator realizes
only a whole number of growth passes, so widths within one class give
identical masks. In
(\ref{eq:general_knob_loss}), $\lambda_j=1$ for the four offsets and
$\lambda_j=4$ for the width class. The layer plan is read from the
logits: for each net and station, the probability mass of the net on
M9 and on M8 is summed over a window around the station, and the layer
with the larger mass is selected. The windows of the two crossovers
are placed at the predicted crossover position.


\begin{thebibliography}{99}
\bibliographystyle{IEEEtran}

\bibitem{bandler2004space}
J. W. Bandler, Q. S. Cheng, S. A. Dakroury, A. S. Mohamed, M. H. Bakr, K. Madsen, and J. S{\o}ndergaard, "Space mapping: The state of the art," \textit{IEEE Trans. Microw. Theory Techn.}, vol. 52, no. 1, pp. 337--361, Jan. 2004.

\bibitem{liu2006integrated}
K. Liu and R. C. Frye, "Full-Circuit Design Optimization of a RF Silicon Integrated Passive Device," \textit{2006 IEEE Electrical Performance of Electronic Packaging, Scottsdale}, AZ, USA, 2006, pp. 327-330.

\bibitem{liu2011synthesis}
B. Liu, D. Zhao, P. Reynaert, and G. G. E. Gielen, "Synthesis of integrated passive components for high-frequency RF ICs based on evolutionary computation and machine learning techniques," \textit{IEEE Trans. Comput.-Aided Design Integr. Circuits Syst.}, vol. 30, no. 10, pp. 1458--1468, Oct. 2011.

\bibitem{gregory2011fast}
M. D. Gregory, Z. Bayraktar and D. H. Werner, "Fast Optimization of Electromagnetic Design Problems Using the Covariance Matrix Adaptation Evolutionary Strategy," in \textit{IEEE Transactions on Antennas and Propagation}, vol. 59, no. 4, pp. 1275-1285, April 2011.

\bibitem{koziel2015rapid}
S. Koziel and P. Kurgan, "Rapid design of miniaturised branch-line couplers through concurrent cell optimisation and surrogate-assisted fine-tuning," \textit{IET Microw. Antennas Propag.}, vol. 9, no. 9, pp. 957--963, 2015.

\bibitem{chenna2026pixelated}
V. Chenna and H. Hashemi, "Algorithmic Design and Optimization of On-Chip Multilayered Pixelated Passive Networks," in \textit{IEEE Transactions on Microwave Theory and Techniques}, vol. 74, no. 1, pp. 356-366, Jan. 2026.


\bibitem{molesky2018inverse}
S. Molesky, Z. Lin, A. Y. Piggott, W. Jin, J. Vu\v{c}kovi\'{c}, and A. W. Rodriguez, "Inverse design in nanophotonics," \textit{Nature Photon.}, vol. 12, no. 11, pp. 659--670, Nov. 2018.

\bibitem{john1996evolutionary}
A. John and R. H. Jansen, "Evolutionary generation of (M)MIC component shapes using 2.5D EM simulation and discrete genetic optimization," in \textit{IEEE MTT-S Int. Microw. Symp. Dig.}, vol. 2, Jun. 1996, pp. 745--748.

\bibitem{jin2005parallel}
N. Jin and Y. Rahmat-Samii, "Parallel particle swarm optimization and finite-difference time-domain (PSO/FDTD) algorithm for multiband and wide-band patch antenna designs," \textit{IEEE Trans. Antennas Propag.}, vol. 53, no. 11, pp. 3459--3468, Nov. 2005.

\bibitem{lee2023pixelated}
J. Lee, W. Jia, B. S. -. Rodriguez and J. S. Walling, "Pixelated RF: Random Metasurface Based Electromagnetic Filters," \textit{2023 21st IEEE Interregional NEWCAS Conference (NEWCAS)}, Edinburgh, United Kingdom, 2023, pp. 1-5.

\bibitem{karahan2024inverse}
Karahan, E.A., Liu, Z., Gupta, A. \textit{et al.} Deep-learning enabled generalized inverse design of multi-port radio-frequency and sub-terahertz passives and integrated circuits. \textit{Nat Commun} 15, 10734 (2024).

\bibitem{hassan2014topology}
E. Hassan, E. Wadbro, and M. Berggren, "Topology optimization of metallic antennas," \textit{IEEE Trans. Antennas Propag.}, vol. 62, no. 5, pp. 2488--2500, May 2014.

\bibitem{vercruysse2019analytical}
D. Vercruysse, N. V. Sapra, L. Su, R. Trivedi, and J. Vu\v{c}kovi\'{c}, "Analytical level set fabrication constraints for inverse design," \textit{Sci. Rep.}, vol. 9, no. 1, p. 8999, Jun. 2019.

\bibitem{capek2023optimal}
M. Capek, L. Jelinek, P. Kadlec, and M. Gustafsson, "Optimal inverse design based on memetic algorithms---Part I: Theory and implementation," in \textit{IEEE Transactions on Antennas and Propagation}, vol. 71, no. 11, pp. 8806--8816, Nov. 2023.

\bibitem{zheng2023ultrafast}
Y. Zheng and C. Sideris, "Ultra-fast Simulation and Inverse Design of Metallic Antennas," \textit{2023 IEEE/MTT-S International Microwave Symposium - IMS 2023}, San Diego, CA, USA, 2023, pp. 351-354.

\bibitem{karahan2023deep}
E. A. Karahan, Z. Liu and K. Sengupta, "Deep-Learning-Based Inverse-Designed Millimeter-Wave Passives and Power Amplifiers," in \textit{IEEE Journal of Solid-State Circuits}, vol. 58, no. 11, pp. 3074-3088, Nov. 2023.

\bibitem{he2025motif}
H. He, Y. Xu, L. Xia, Y. Hu, F. Cai and T. Chi, "MOTIF-RF: Multi-template On-chip Transformer Synthesis Incorporating Frequency-domain Self-transfer Learning for RFIC Design Automation," \textit{2026 31st Asia and South Pacific Design Automation Conference (ASP-DAC)}, Lantau, Hong Kong, 2026, pp. 1138-1144.

\bibitem{wei2023transformer}
Y. Wei, J. Wei, Q. Wu and H. Wang, "Machine Learning-Assisted On-chip Quadrilateral Interleaved Transformer Automatic Synthesis," \textit{2023 16th UK-Europe-China Workshop on Millimetre Waves and Terahertz Technologies (UCMMT)}, Guangzhou, China, 2023, pp. 1-3.


\bibitem{chae2024pulserf}
H. Chae, H. Yu, S. Li and D. Z. Pan, "PulseRF: Physics-Augmented ML Modeling and Synthesis for High-Frequency RFIC Design," \textit{2024 ACM/IEEE International Conference On Computer Aided Design (ICCAD)}, Newark, NJ, USA, 2024, pp. 01-09.

\bibitem{wolpert1997no}
D. H. Wolpert and W. G. Macready, "No free lunch theorems for optimization," in \textit{IEEE Transactions on Evolutionary Computation}, vol. 1, no. 1, pp. 67-82, April 1997.

\bibitem{hinton2006reducing}
G. E. Hinton and R. R. Salakhutdinov, "Reducing the dimensionality of data with neural networks," \textit{Science}, vol. 313, no. 5786, pp. 504--507, Jul. 2006.

\bibitem{gomez2018automatic}
R. G\'{o}mez-Bombarelli \textit{et al.}, "Automatic chemical design using a data-driven continuous representation of molecules," \textit{ACS Central Sci.}, vol. 4, no. 2, pp. 268--276, 2018.

\bibitem{tripp2020sample}
A. Tripp, E. Daxberger, and J. M. Hern\'{a}ndez-Lobato, "Sample-efficient optimization in the latent space of deep generative models via weighted retraining," in \textit{Proceedings of the 2020 Advances in Neural Information Processing Systems (NeurIPS)}, vol. 33, 2020, pp. 11259--11272.

\bibitem{hart1968formal}
P. E. Hart, N. J. Nilsson, and B. Raphael, "A formal basis for the heuristic determination of minimum cost paths," \textit{IEEE Transactions on Systems Science and Cybernetics}, vol. 4, no. 2, pp. 100--107, Jul. 1968.

\bibitem{nelder1965simplex}
J. A. Nelder and R. Mead, "A simplex method for function minimization," \textit{The Computer Journal}, vol. 7, no. 4, pp. 308--313, 1965.

\bibitem{hansen2001completely}
N. Hansen and A. Ostermeier, "Completely Derandomized Self-Adaptation in Evolution Strategies," in \textit{Evolutionary Computation}, vol. 9, no. 2, pp. 159-195, June 2001.


\bibitem{hansen2016cma}
N. Hansen, "The CMA evolution strategy: A tutorial," 2016, \textit{arXiv:1604.00772}.

\bibitem{auger2005restart}
A. Auger and N. Hansen, "A restart CMA evolution strategy with increasing population size," in \textit{Proceedings of the 2005 IEEE Congress on Evolutionary Computation (CEC)}, 2005, pp. 1769--1776.

\bibitem{evensen1994sequential}
G. Evensen, "Sequential data assimilation with a nonlinear quasi-geostrophic model using Monte Carlo methods to forecast error statistics," \textit{Journal of Geophysical Research: Oceans}, vol. 99, no. C5, pp. 10143--10162, 1994.

\bibitem{iglesias2013ensemble}
M. A. Iglesias, K. J. H. Law, and A. M. Stuart, "Ensemble Kalman methods for inverse problems," \textit{Inverse Problems}, vol. 29, no. 4, p. 045001, 2013.

\bibitem{zheng2024ensemble}
H. Zheng, W. Chu, A. Wang, N. Kovachki, R. Baptista, and Y. Yue, "Ensemble Kalman diffusion guidance: A derivative-free method for inverse problems," 2024, \textit{arXiv:2409.20175}.

\bibitem{Zolkov2026}
E. Zolkov, Y. Ginzberg, N. Ginzberg and E. Cohen, "A 65-nm CMOS mm-Wave Blocker-Tolerant Digital Receiver Array," in \textit{IEEE Transactions on Microwave Theory and Techniques}, vol. 74, no. 2, pp. 1981-1993, Feb. 2026.

\bibitem{chenna2026inverse}
V. Chenna and H. Hashemi, "Inverse Design of Multilayered Pixelated mm-Wave Power Amplifiers," in \textit{IEEE Journal of Solid-State Circuits}, vol. 61, no. 5, pp. 1764-1775, May 2026.


\bibitem{Garay}
E. Garay, M.-Y. Huang, and H. Wang, "A cascaded self-similar rat-race hybrid coupler architecture and its compact fully integrated Ka-band implementation," in \textit{IEEE MTT-S International Microwave Symposium Digest}, Philadelphia, PA, USA, 2018, pp. 79--82.

\bibitem{Parveg}
D. Parveg, M. Varonen, D. Karaca, and K. Halonen, "Wideband mm-wave CMOS slow wave coupler," \textit{IEEE Microwave and Wireless Components Letters}, vol. 29, no. 3, pp. 210--212, Mar. 2019.

\bibitem{Acri}
G. Acri \textit{et al.}, "A millimeter-wave miniature branch-line coupler in 22-nm CMOS technology," \textit{IEEE Solid-State Circuits Lett.}, vol. 2, no. 6, pp. 45--48, Jun. 2019.

\bibitem{Qiu}
F. Qiu, H. Zhu, L. Wu, W. Che, and Q. Xue, "A 15--38 GHz vector-summing phase-shifter with 360$^\circ$ phase-shifting range using improved I/Q generator," \textit{IEEE Transactions on Circuits and Systems II: Express Briefs}, vol. 68, no. 10, pp. 3199--3203, Oct. 2021.

\bibitem{vincent2008extracting}
P. Vincent, H. Larochelle, Y. Bengio, and P.-A. Manzagol, "Extracting and composing robust features with denoising autoencoders," in \textit{Proceedings of the 25th international conference on Machine learning (ICML '08)}, 2008, pp. 1096--1103.

\bibitem{kingma2014auto}
D. P. Kingma and M. Welling, "Auto-encoding variational Bayes," in \textit{Proceedings of the 2nd International Conference on Learning Representations (ICLR)}, 2014.

\bibitem{szegedy2015going}
C. Szegedy \textit{et al.}, "Going deeper with convolutions," in \textit{Proceedings of the 2015 IEEE Conference on Computer Vision and Pattern Recognition (CVPR)}, 2015, pp. 1--9.

\bibitem{he2016deep}
K. He, X. Zhang, S. Ren, and J. Sun, "Deep residual learning for image recognition," in \textit{Proceedings of the 2016 IEEE Conference on Computer Vision and Pattern Recognition (CVPR)}, 2016, pp. 770--778.

\bibitem{wu2018group}
Y. Wu and K. He, "Group normalization," in \textit{Proceedings of the 2018 European Conference on Computer Vision (ECCV)}, 2018, pp. 3--19.

\bibitem{liu2018intriguing}
R. Liu \textit{et al.}, "An intriguing failing of convolutional neural networks and the CoordConv solution," in \textit{Proceedings of the 2018 Advances in Neural Information Processing Systems (NeurIPS)}, vol. 31, 2018.

\bibitem{shi2016real}
W. Shi \textit{et al.}, "Real-time single image and video super-resolution using an efficient sub-pixel convolutional neural network," in \textit{Proceedings of the 2016 IEEE Conference on Computer Vision and Pattern Recognition (CVPR)}, Las Vegas, NV, USA, 2016, pp. 1874-1883.

\end{thebibliography}
\end{document}